\documentclass[journal]{IEEEtran}
\ifCLASSINFOpdf
  \usepackage[pdftex]{graphicx}
  \usepackage[noadjust]{cite}
  \usepackage{color}

  \usepackage{amsmath}
\else
\fi

\begin{document}

\newcommand{\red}{\color[rgb]{0.6,0.0,0.0}}
\newcommand{\yellow}{\color[rgb]{0.6,0.6,0.0}}
\newcommand{\green}{\color[rgb]{0.0,0.5,0.0}}
\newcommand{\blue}{\color[rgb]{0.0,0.4,0.6}}
\newcommand{\purple}{\color[rgb]{0.6,0.15,0.7}}

\newcommand{\josh}[1]{{[\bf \blue {Josh: } #1}]}
\newcommand{\hc}[1]{{[\bf \green {Henry C: } #1}]}
\newcommand{\ck}[1]{{[\bf \red {Charlie: } #1}]}
\newcommand{\tapo}[1]{{[\bf \purple {Tapo: } #1}]}

%

\title{Material Recognition via Heat Transfer Given Ambiguous Initial Conditions}




%
%
%

 \author{Tapomayukh Bhattacharjee*,~\IEEEmembership{Student Member,~IEEE,} Henry M.~Clever*$^{\dagger}$,~\IEEEmembership{Student Member,~IEEE,} Joshua~Wade,~\IEEEmembership{Student Member,~IEEE,} Charles C.~Kemp,~\IEEEmembership{Member,~IEEE}
\thanks{All authors were with the Department
of Department of Biomedical Engineering, Georgia Institute of Technology, Atlanta,
GA, 30332 USA.}
\thanks{*Authors contributed equally.} \thanks{$^{\dagger}$Corresponding author, email: henryclever@gatech.edu.  }
\thanks{Manuscript received October XX 2020; revised XXXXX.}}

\markboth{Journal of \LaTeX\ Class Files,~Vol.~14, No.~8, August~2015}%
{Shell \MakeLowercase{\textit{et al.}}: Bare Demo of IEEEtran.cls for IEEE Journals}
%

\maketitle

\begin{abstract}
Humans and robots can recognize materials with distinct thermal effusivities by making physical contact and observing temperatures during heat transfer. This works well with room temperature materials and humans and robots at human body temperatures. Past research has shown that cooling or heating a material can result in temperatures that are similar to contact with another material. To thoroughly investigate this perceptual ambiguity, we designed a psychophysical experiment in which a participant discriminates between two materials given ambiguous initial conditions. We conducted a study with 32 human participants and a robot. Humans and the robot confused the materials. We also found that robots can overcome this ambiguity using two temperature sensors with different temperatures prior to contact. We support this conclusion based on a mathematical proof using a heat transfer model and empirical results in which a robot achieved $100\%$ accuracy compared to $5\%$ human accuracy. Our results also indicate that robots can use subtle cues to distinguish thermally ambiguous materials with a single temperature sensor. Overall, our work provides insights into challenging conditions for material recognition via heat transfer, and suggests methods by which robots can overcome these challenges to outperform humans. 
\end{abstract}


%
\IEEEpeerreviewmaketitle


\section{Introduction}
%
%
%
%

\IEEEPARstart{T}{he} sense of touch enables humans to dexterously manipulate objects under diverse conditions, including situations with visual occlusion such as reaching into foliage or a cluttered refrigerator \cite{jain2013reaching}. Upon touching an object, heat transfers between the skin and the object in a manner that depends on the properties of the object. Humans can use cutaneous thermoreceptors to sense this heat transfer and gain information \cite{katz1925world, dyck1974description, tritsch1988veridical, jones2003material, ho2004material, ho2006contribution,  tiest2010tactual, tiest2009tactile, ho2018material}. Similarly, robots can gain information by making contact with a heated probe that includes temperature sensors \cite{kron2003multi, bhattacharjee2015material, bhattacharjee2016data, katoh2009material, engel2006flexible, chu2015robotic, takamuku2008robust, kerr2013material}. 

Researchers have usefully modeled this form of sensing as heat transfer between two semi-infinite solids \cite{bhattacharjee2015material,ho2006contribution,jones2008warm,yamamoto2004control}. In this paper, we refer to this model as the \textit{simplified heat transfer model}. For this mathematical model, the thermal effusivity and initial temperature of the material being contacted play important roles. Notably, the model predicts that materials with distinct effusivities can result in the same measured temperatures over time. For example, contact with ambient temperature metal could result in the same temperatures over time as contact with cold wood. This has inspired researchers to create thermal haptic displays that actively alter the temperature of a material to simulate the sensation of touching other materials \cite{guiatni2008theoretical,jones2008warm,yang2009spatial,ino1993tactile,yamamoto2004control,ho2004material}. In this paper, we refer to initial conditions that the simplified heat transfer model predicts would result in the same measured temperatures as \textit{ambiguous conditions}. 

\begin{figure}[!t]
\begin{center}
\includegraphics[width=8.7cm]{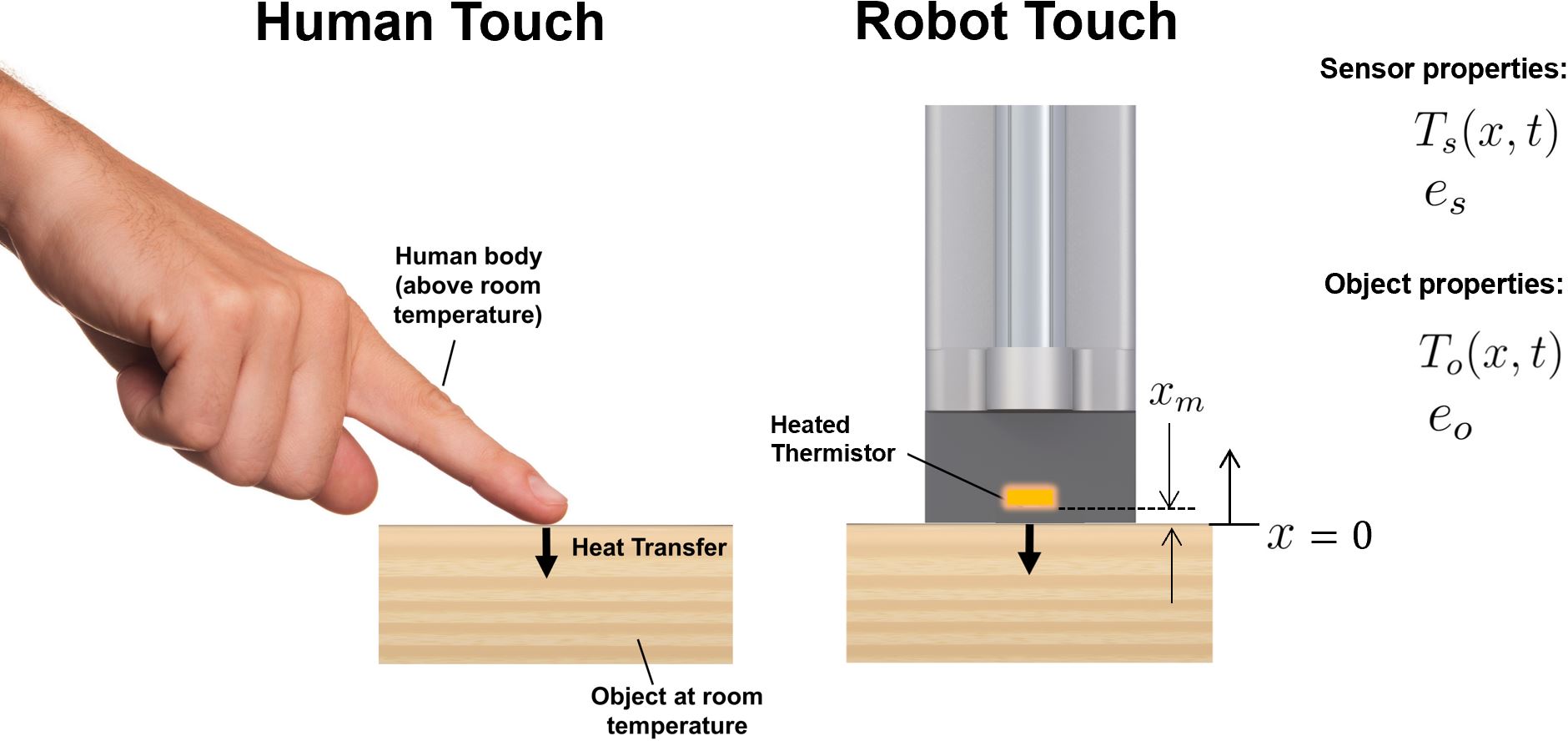} 
\vspace{-2mm}
\caption{\label{fig:1}\textbf{Modeling Thermal Perception.} When the human body makes contact with an object at ambient temperature, heat flows from the body into the object, creating the sensation of coldness that is more or less pronounced based on the thermal properties of the object. Similarly a robot with an actively heated thermal sensor can use this sensing modality to distinguish materials with different thermal properties. We model the heat transfer from the robot's sensor to the object as contact between two semi-infinite solids, an approach also used to approximate human thermal touch  \cite{bhattacharjee2015material,ho2006contribution,jones2008warm,yamamoto2004control}.  The simplified heat transfer model depends on material and sensor initial temperature $T$ and effusivity $e$. The ability to control the robot temperature enables the robot to overcome human limits of thermal perception.
}
\end{center}
\end{figure}

Our main motivation is to enable robots to better perceive the world through touch. With respect to this objective, ambiguous conditions could limit robotic performance and represent a challenge to overcome. To thoroughly investigate the type of perceptual ambiguity predicted by the simplified heat transfer model, we designed a psychophysical experiment that both humans and robots can perform. The experiment assesses how well a participant can discriminate between two materials given ambiguous conditions. We used two materials commonly found in human environments with which human participants have extensive experience, metal and wood. We also restricted humans and robots to perform a simple touching motion and deprived human participants of visual and auditory information about the materials. 

To characterize human performance, we conducted a study with 32 human participants. We found that ambiguous conditions dramatically reduced human performance. For trials with ambiguous conditions, participants misidentified cold wood as metal in 93.8\% of the trials. In contrast, they correctly identified ambient temperature wood as wood in 100\% of the trials and ambient temperature metal as metal in 94\% of the trials. Interestingly, these errors are consistent with the assumption that the contacted material was at the ambient temperature. For the simple touching motion and the smooth material samples, our results suggest that heat transfer dominated the decision and that other touch modalities played insignificant roles. 

We used the same psychophysical experiment for three different robotic investigations. For the first robotic investigation, we created a robot model of human sensing that performed similarly to the human participants. For this robot, we trained a classifier on data from ambient temperature wood and metal. For the second robotic investigation, we trained a classifier on data from cold wood as well as ambient temperature wood and metal. With this additional training data, the robot correctly classified thermally ambiguous materials 79\% of the time. This suggests that the robot used subtle differences between the real signals and the idealized signals predicted by the simplified heat transfer model to overcome the ambiguous conditions. Through a post-hoc analysis, we identified the slope of the temperature signals near the start of contact as candidate features for successful classification. An open question is whether humans could achieve comparable performance through additional training. 

For the third robotic investigation, the robot used a sensor with two temperature sensors at different initial temperatures, which is distinct from human sensing. We mathematically show that this sensor design eliminates the ambiguity predicted by the simplified heat transfer model. Empirically, the robot achieved perfect performance, indicating that this approach overcomes the ambiguity in practice and can enable robots to outperform humans.

\section{Related Work}

Our research on material recognition given ambiguous conditions for heat transfer relates to prior work on human and robot perception. 

\subsection{Human Thermal Perception}

Early work in 1925 by Katz~\cite{katz1925world} explored the mechanisms by which humans perceive the world through touch, such as how thermal cues can be used to recognize materials. Dyck \textit{et al.}~\cite{dyck1974description} found that humans could reliably distinguish metal and plastic as ``cold'' and ``warm'' materials at ambient temperatures. Many works have since studied the influence of thermal cues on material recognition \cite{tritsch1988veridical, tiest2009tactile,ho2004material, ho2006contribution, jones2003material, tiest2010tactual, ho2018material}; for example, Bergmann Tiest and Kappers~\cite{tiest2009tactile} studied how humans can discriminate materials from heat extraction rate, i.e. which one cools faster. Some have explored the effect of contact area on thermal perception~\cite{jones2008warm}, and Yang \textit{et al.}~\cite{yang2008use} show how using multiple fingers can improve thermal perception. 

Heat transfer depends on various factors, including the initial temperatures and material properties of the object and contacting body, as well as contact properties such as pressure and the geometry of contact \cite{weber1978eh, hensel1973cutaneous, tiest2007experimentally, burnett1927experience, stevens1960warmth, tritsch1988veridical, tiest2008thermosensory, green2004temperature}. Some characteristics of thermal perception can lead to confusion when recognizing materials. Tritsch~\cite{tritsch1988veridical} found that human perception is more influenced by the temperature difference between the skin and an object, rather than the object's absolute temperature: whether or not one just returned indoors on a cold day may cause them to perceive objects differently. Moreover, when the skin temperature is very close to the touched object temperature, little information is available for material recognition~\cite{ho2018material}. However, previous work has not rigorously studied the implications of ambiguous conditions on touch perception. 

Researchers have previously modeled heat transfer based interaction between a human finger and an object as contact between two semi-infinite solids \cite{bhattacharjee2015material,ho2006contribution,jones2008warm,yamamoto2004control}. Such a model may be used to predict a change in skin or object temperature when contact is made~\cite{ho2006contribution}, or simulate material perception on thermal displays \cite{ho2004material, yamamoto2004control, jones2008warm}. Our work continues to explore how humans perceive the world through touch, by using this model to create ambiguous conditions between distinct materials and testing the human ability to recognize them.


\subsection{Robot Thermal Perception}


Robots have been shown to benefit from thermal touch sensing, but results have yet to show how thermally ambiguous materials can be distinguished \cite{takamuku2008robust, kerr2013material, xu2013tactile, bhattacharjee2015material, wade2017force}. For a robot, distinguishing object materials can enhance manipulation capabilities and physical interaction with the world. A variety of sensing designs may be used for thermal material perception, including flexible polymer substrates~\cite{engel2006flexible} and heat flow sensing~\cite{katoh2009material}. Takamuku \textit{et al.}~\cite{takamuku2008robust} mounted thermistors in a soft anthropomorphic robot fingertip, and Xu \textit{et al.}~\cite{xu2013tactile} used a BioTac sensor on a robot to identify objects using compliance, texture, and thermal properties. Others have extended thermal sensing over a larger surface area of robots in the form of robotic skin~\cite{wade2017force}. The presence of thermal sensing in robotic skin has been shown to increase the accuracy of material recognition for common household objects  \cite{bhattacharjee2018multimodal}.

In prior research, we equipped a robot with a thermal sensor with double initial conditions ~\cite{wade2017force}. This is similar to a sensor we created for the current paper. The previous sensor was intended to recognize contact with ambient temperature materials and warm human skin. The notion was that an unheated sensor would result in greater heat transfer with human skin than a warm sensor. We also used a similar sensor on a robot that touched objects found in a real home, including objects in a refrigerator ~\cite{bhattacharjee2018multimodal}. However, the papers did not explicitly consider thermally ambiguous conditions nor did they provide theoretical justification for the design.

Prior works have used data-driven models to train robots how to classify common materials from thermal sensors~\cite{kerr2013material,bhattacharjee2016data, wade2017force, bhattacharjee2018multimodal}. Kerr \textit{et al.}~\cite{kerr2013material} used data-driven methods to train a robot to recognize materials using thermal information from the BioTac sensor, and found that the robot outperformed human participants. For the studies in the current paper, we used similar data-driven models, yet unlike prior work we focus on ambiguous conditions.



\section{Mathematical Models}

We use a mathematical model of heat transfer from the literature  to provide insights into the recognition of an object's material by humans and robots \cite{bhattacharjee2015material,ho2006contribution,jones2008warm,yamamoto2004control}. Given various physical parameters, the model predicts temperatures that occur over time during touch perception. Based on this model, we define ambiguous conditions and present a novel proof of the potential for a robotic sensor to overcome ambiguity. 

\subsection{The Simplified Heat Transfer Model}

The simplified heat transfer model depends on the materials' thermal effusivities and their temperatures prior to contact. For heat transfer to occur, the sensor and the object must have different initial temperatures. When human skin, typically warmer than ambient temperature, comes in contact with an object at ambient temperature, heat flows out of the skin at a rate that depends on the initial contact conditions as well as material properties of the skin and the object (See Fig. \ref{fig:1}). A similar phenomena occurs with an actively heated robot sensing module, also shown in Fig. \ref{fig:1}. Given consistent initial conditions, the heat-transfer rate will differ only according to the material properties. As such, data-driven or model-based methods can be used to identify different materials from time series of temperature measurements \cite{bhattacharjee2015material}. 

The model makes a number of simplifying assumptions. It uses a semi-infinite solid model for the sensor and the object being touched and is valid for a short time duration \cite{yang2008use}. It does not consider heat transfer among the air, the object, and the sensor and ignores continued heating of the sensor during contact. It also neglects details of contact, such as pressure and contact area. 



Let $T_{\mathrm{s}}(x,t)$ and $T_{\mathrm{o}}(x,t)$ be the temperatures of the sensor and the object semi-infinite solid models at a distance $x$ from the contact surface and at a time $t$ respectively. Contact between the sensor and the object occurs at $t=0$, and the contact occurs across a plane at $x=0$. Positive values of $x$ represent the distance from a point within the sensor to this plane of contact, while negative values represent the distance from a point within the object to this plane (see Fig.~\ref{fig:1}). 

A sensing element  measures temperatures over time within the sensor at $x=x_m$. We are interested in the potential for a human or robot to infer the object's material based on sensed temperatures over some finite period of time, $T_m(t) = T_{\mathrm{s}}(x=x_m, t)$ where $0 \leq t \leq t_{max}$. We assume that recognition will be performed over a set of materials, $M$, with distinct thermal effusivities, $M=\{e_1, e_2, \ldots, e_n\}$. As such, for our material recognition model, estimation of the object's thermal effusivity, $e_{\mathrm{o}}$, is equivalent to recognition of the object's material. The thermal effusivity of a material, $e$, is a function of the thermal conductivity, $\lambda$, the volumetric mass density, $\rho$, and the specific heat capacity, $c_p$, such that $e=\sqrt{\lambda \rho c_p}$ \cite{cengel2014heat}.

\begin{figure*}[!t]
\begin{center}
\includegraphics[width=18cm]{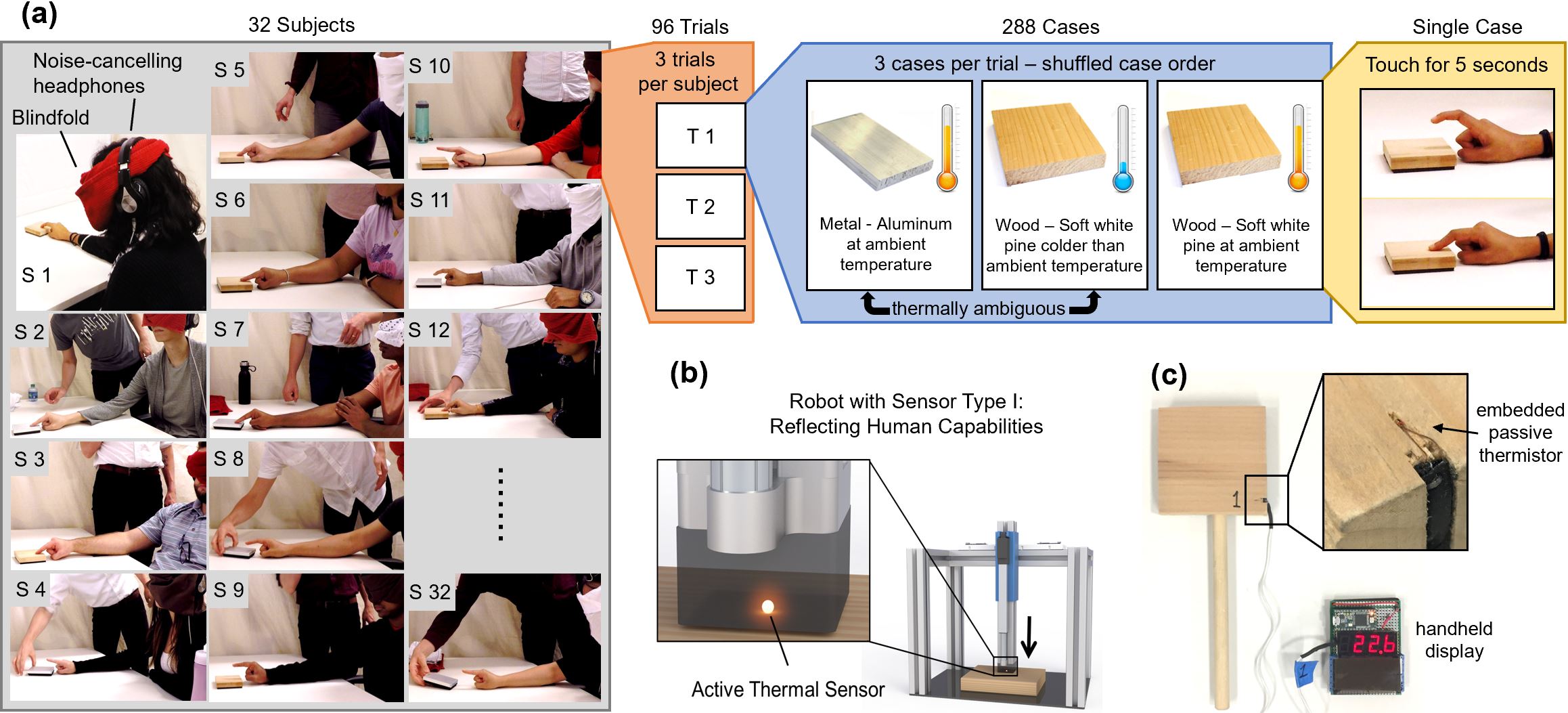} 
\vspace{-2mm}
\caption{\label{fig:2}\textbf{Psychophysical Experiment.} (a) Humans use touch sensing with a single finger to recognize materials at ambiguous conditions. We use a forced-choice clustered-paired experiment to test if aluminum and cooled wood feel similar and are thus confused. 
(b) We used a simple 1-DoF robot that models human thermal sensing capabilities to test samples using the same sample regimen as the human study ($N=30$ trial sets). The robot lowers the linear actuator to bring the sensor module into contact with the material sample and records temperature values from a thermistor. (c) Cold wood temperature was monitored and recorded at the time it was touched, using a thermistor embedded into the surface of the cold wood.
}
\end{center}
\end{figure*}

We assume that the semi-infinite solids for the sensor and the object have constant, uniform temperatures prior to contact, $\mathcal{T}_{\mathrm{s}} = T_{\mathrm{s}}(x,t<0)$ and  $\mathcal{T}_{\mathrm{o}} = T_{\mathrm{o}}(x, t<0)$. Instantaneously upon contact, the contact surface attains the temperature $\mathcal{T}_{\mathrm{c}} = T_{\mathrm{s}}(x=0,t=0) = T_{\mathrm{o}}(x=0, t=0)$. 

Given our assumptions, the measured temperatures over time can be expressed as 
\begin{equation}\label{eq:1}
{T_m\left( {t} \right) = \mathcal{T}_{\mathrm{s}} + \left(\mathcal{T}_{\mathrm{c}} - \mathcal{T}_{\mathrm{s}}\right) * \mathrm{erfc}\left( {\frac{x_m}
{{2\sqrt {{\alpha _\mathrm{s}}t} }}} \right)}
\end{equation}
where $\mathrm{erfc}$ is the complementary error function and $\alpha_{\mathrm{s}}$ is the coefficient of thermal diffusivity of the sensor's material. We can further express the initial contact surface temperature as \begin{equation}\label{eq:2}
\mathcal{T}_{\mathrm{c}} = \frac{{\mathcal{T}_{\mathrm{s}} e_{\mathrm{s}}} + {\mathcal{T}_{\mathrm{o}} e_{\mathrm{o}}}} {e_{\mathrm{s}} + e_{\mathrm{o}}}
\end{equation}
where $e_{\mathrm{s}}$ is the thermal effusivity of the sensor's material and $e_{\mathrm{o}}$ is the thermal effusivity of the object's material. 


%

If the measured temperatures, $T_m\left( {t} \right)$ for $0\leq t \leq t_{max}$, are unique for each of the materials in $M$, then under ideal circumstances the materials can be successfully recognized. In practice, the material effusivities, $M=\{e_1, e_2, \ldots, e_n\}$, must result in measured temperatures that are sufficiently different from one another that they can be distinguished given factors such as noise and the rate of measurements. When all other parameters are held constant in eq. (\ref{eq:1}) and only $e_{\mathrm{o}}$ is allowed to vary, the measured temperatures are unique for distinct values of $e_{\mathrm{o}}$. By inspection, one can see that distinct values of $e_{\mathrm{o}}$ result in distinct values for $\mathcal{T}_{\mathrm{c}}$ and hence unique functions for $T_m\left( {t} \right)$.

\subsection{Ambiguous Conditions}

When the initial temperature of the object, $\mathcal{T}_{\mathrm{o}}$, is also allowed to vary, distinct values of $e_{\mathrm{o}}$ may not result in distinct measurements. In this manuscript, we use our mathematical model to focus on this class of ambiguous situations.

We can consider the minimal material recognition problem to be a set of materials, $M_{min}$, with two objects that have distinct thermal effusivities, $M_{min} = \{e_{\mathrm{o1}}, e_{\mathrm{o2}}\}$. An ambiguous situation occurs when temperature measurements from contact with the two objects are the same, so $T_{m1}\left( {t} \right) = T_{m2}\left( {t} \right)$ for $0\leq t \leq t_{max}$ where $T_{m1}$ and $T_{m2}$ represent temperature measurements from contact with objects 1 and 2, respectively. Similarly, we can define $\mathcal{T}_{c1}$ and $\mathcal{T}_{c2}$ to be the initial contact surface temperatures for objects 1 and 2. As shown in eq. (\ref{eq:2}), the initial contact surface temperature depends on both the object's thermal effusivity and the initial temperature of the object. 

Using eq. (\ref{eq:1}), we can determine the conditions that the simplified heat transfer model predicts will lead to ambiguity. 

\begin{equation}\label{eq:3}
\begin{split}
T_{m1}\left( {t} \right) &= T_{m2}\left( {t} \right) \\
\mathcal{T}_{\mathrm{s}} + \left(\mathcal{T}_{c1} - \mathcal{T}_{\mathrm{s}}\right) {\scriptstyle * \mathrm{erfc}\left( {\frac{x_m} {{2\sqrt {{\alpha _\mathrm{s}}t} }}} \right)} &= \mathcal{T}_{\mathrm{s}} + \left(\mathcal{T}_{c2} - \mathcal{T}_{\mathrm{s}}\right) {\scriptstyle * \mathrm{erfc}\left( {\frac{x_m} {{2\sqrt {{\alpha _\mathrm{s}}t} }}} \right)} \\
\mathcal{T}_{c1} &= \mathcal{T}_{c2} \\
\frac{{\mathcal{T}_{\mathrm{s}} e_{\mathrm{s}}} + {\mathcal{T}_{\mathrm{o1}} e_{\mathrm{o1}}}} {e_{\mathrm{s}} + e_{\mathrm{o1}}} &= \frac{{\mathcal{T}_{\mathrm{s}} e_{\mathrm{s}}} + {\mathcal{T}_{\mathrm{o2}} e_{\mathrm{o2}}}} {e_{\mathrm{s}} + e_{\mathrm{o2}}}
\end{split}
\end{equation}

We can now solve for the initial temperature of object 2, $\mathcal{T}_{\mathrm{o2}}$, that would result in measurements identical to those obtained from object 1. We used an algebraic solver (WolframAlpha online \cite{wolframalpha}) to find the following equation for the ambiguous object temperature for object 2,

\begin{equation}\label{eq:4}
\mathcal{T}_{\mathrm{o2}} = \frac{-\mathcal{T}_{\mathrm{s}} e_{\mathrm{s}} e_{\mathrm{o1}} + \mathcal{T}_{\mathrm{s}} e_{\mathrm{s}} e_{\mathrm{o2}} + e_{\mathrm{s}} \mathcal{T}_{\mathrm{o1}} e_{\mathrm{o1}} + \mathcal{T}_{\mathrm{o1}} e_{\mathrm{o1}} e_{\mathrm{o2}}}{e_{\mathrm{o2}} (e_{\mathrm{s}} + e_{\mathrm{o1}})}
\end{equation}

which holds due to $e_{\mathrm{o2}} \neq 0$, $e_{\mathrm{s}} + e_{\mathrm{o1}} \neq 0$, and $e_{\mathrm{s}} + e_{\mathrm{o2}} \neq 0$, since all thermal effusivities are strictly greater than 0. 


\subsection{Overcoming Ambiguous Conditions}

We can prove that simultaneously using two temperature sensors with distinct initial temperatures overcomes this ambiguity. We refer to this type of sensor as a \emph{double-condition sensor}. 

Let the initial temperatures of the two sensors be $\mathcal{T}_{\mathrm{s1}}$ and $\mathcal{T}_{\mathrm{s2}}$, with effusivities $e_{\mathrm{s1}}$ and $e_{\mathrm{s2}}$. Substituting these in Eq. (\ref{eq:4}) results in a distinct ambiguous temperature for object 2, $\mathcal{T}_{\mathrm{o2}}$ per sensor, only if
\begin{equation}\label{eq:5}
\begin{split}
\Big(\mathcal{T}_{\mathrm{s1}}e_{\mathrm{s1}}\big(e_{\mathrm{s2}} + e_{\mathrm{o1}}\big) + \mathcal{T}_{\mathrm{o1}}e_{\mathrm{o1}}\big(e_{\mathrm{s2}} - e_{\mathrm{s1}}\big)\Big)\big(e_{\mathrm{o2}}  - e_{\mathrm{o1}} \big) \neq  \\ \mathcal{T}_{\mathrm{s2}}e_{\mathrm{s2}}\big(e_{\mathrm{s1}} + e_{\mathrm{o1}}\big)\big(e_{\mathrm{o2}} - e_{\mathrm{o1}}\big) \\
\end{split}
\end{equation}
which holds if $e_{\mathrm{o2}} \neq e_{\mathrm{o1}}$.  Appendix A has the detailed proof. 
Consequently, the model predicts that only one of the two temperature sensors can produce ambiguous measurements for the two objects; one of the the two sensors will always produce unambiguous measurements. 

\begin{figure*}
\begin{center}
\includegraphics[width=18cm]{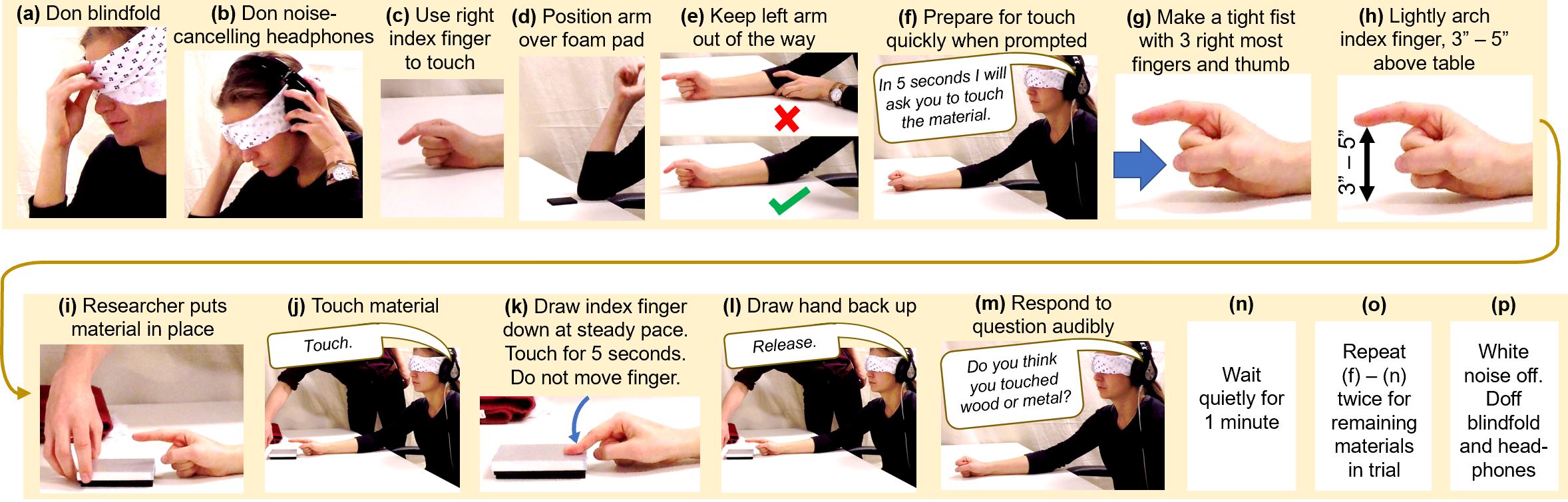} 
\vspace{-2mm}
\caption{\label{fig:3}\textbf{Human Study Protocol - Single Trial Sequence.}   Temporally sensitive portions of the trial were prompted verbally over the headphones; others, such as how to make the appropriate fist (g), were instructed prior to the beginning of the trial. There were no accidents (i.e. touching a material with two fingers) in the study. All trials were video recorded except one subject who declined to be recorded. $(3$ trials per subject$)*(32$ subjects$) = 96$ trials were conducted, for a total of $(96$ trials$)*(3$ cases per trial$) = 288$ material block touch cases.}
\end{center}
\end{figure*}

\section{The Psychophysical Experiment}


We designed a novel psychophysical experiment to assess how well people and robots can discriminate materials given ambiguous conditions. 

\subsection{Overview of the Experiment}

We created ambiguous conditions with an aluminum block ($o1$) and a soft white pine wood block ($o2$). Metal and wood are common materials found within human environments that are generally considered to feel different when touched. We used controllable refrigeration equipment to adjust the temperature of the pine wood, $\mathcal{T}_{\mathrm{o2}}$, close to the value that eq. (\ref{eq:4}) predicted would result in perceptual ambiguity with ambient temperature aluminum, $\mathcal{T}_{\mathrm{o1}}$. 

We conducted an experiment with $N=32$ human participants and presented the robot with the same sample regimen (see Fig.~\ref{fig:2}). Each participant was given three trials. In each of the three trials, the experimenter placed three material samples in front of the participant in a random order: ambient temperature metal, ambient temperature wood, and cold wood. The participant would then touch the material sample and state whether it was metal or wood (i.e., a forced-choice design).

\subsection{Preparation of Ambiguous Material Samples}

To achieve the thermally ambiguous temperature condition for the pine wood material sample, we used a SO-LOW CH25-3 controllable deep freezer and a SO-LOW RIN10-20 refrigerated incubator to cover temperature ranges of $-35^\circ C$ to $0^\circ C$ and $0^\circ C$ to $22.5^\circ C$, respectively. This range was sufficient to create an ambiguous condition of cold wood for all participants in the study. Participants had a min and max finger temperature of $24.6^\circ C$ and  $32.9^\circ C$, requiring the wood to be cooled to a range between $-35^\circ C$ and  $8^\circ C$, which accounts for pre-touch material warming.  Both the freezer and refrigerated incubator were kept behind a curtain that divided the room so that subjects were unaware of them.

We embedded a temperature sensor in the surface of the wooden block to measure its temperature at the time the participant touched it, as shown in Fig. \ref{fig:2}(c). To facilitate smooth egress from the refrigerator, the cold wood blocks were prepared using the handle shown.  

We assumed that the remaining ambient temperature material blocks did not deviate significantly and therefore we did not monitor them. We did not use a handle on ambient temperature blocks because without having to go behind a curtain and maneuver a refrigerator, preparation was faster and easier.

We obtained the thermal effusivity of the pine wood block ($e_\mathrm{o1} = 331~ \mathrm{J/(m^2s^{0.5}K)}$, the aluminum block ($e_\mathrm{o2} = 23664~ \mathrm{J/(m^2s^{0.5}K)}$), and the robot sensor ($e_\mathrm{r} = 892~ \mathrm{J/(m^2s^{0.5}K)}$) by using a quasi-Newton constrained nonlinear optimization method on empirical data collected by the robot shown in Fig.~\ref{fig:2}(b). For this, the material blocks were initially at $\mathcal{T}_\mathrm{o1} = 22.5^\circ$C and the robot sensor was at $\mathcal{T}_\mathrm{r} =35.0^\circ$C. We used a finger thermal effusivity of $e_\mathrm{f} = 1450 W \cdot s^{1\over 2}·m^{-2}·K^{-1}$ \cite{yoshida2010measurement}.

While piloting our experiment, we found that frost can alter perceived material properties. We took steps to reduce the likelihood of frost, including reducing the time between refrigeration and sensor contact, and using multiple refrigerated wood blocks to avoid frost buildup. 

\section{The Human Study}\label{sec:human_study}

We conducted a human study with approval from the Georgia Institute of Technology Institutional Review Board (IRB) and obtained informed consent from all participants. During our study, participants donned a blindfold and noise cancelling headphones which played back pre-recorded prompts for using their right index finger to touch a material for 5 seconds, shown in Fig. \ref{fig:3}. After the participant moved their finger back up, they were asked ``\textit{Do you think you touched wood or metal?}'' 

\subsection{Protocol}

\subsubsection{Skin Temperature}

The temperature of a person's skin, $\mathcal{T}_{\mathrm{f}}$, plays a key role in the ability to gather thermal information. When we apply the simplified heat transfer model to human perception, the sensor temperature, $\mathcal{T}_{\mathrm{s}}$, is equal to the person's finger temperature, $\mathcal{T}_{\mathrm{s}}=\mathcal{T}_{\mathrm{f}}$.  Human finger temperature can vary significantly among individuals due to factors such as capillary flow \cite{rubenstein1990skin}. The perceptual ambiguity we studied depends on finger temperature, so we measured participants' finger temperatures. Since it is challenging to rapidly change the temperature of a material, our experiment consisted of two sessions. For the first, we measured finger temperature, $\mathcal{T}_{\mathrm{f}}$, and used $\mathcal{T}_{\mathrm{s}} = \mathcal{T}_{\mathrm{f}}$ in the analytical model (i.e. Eq. 4) to predict the conditions for ambiguity. For the second, we refrigerated the wood based to the predicted ambiguous temperature $T_{\mathrm{o2}}$ and tested the person's ability to distinguish the cold wood from the metal.

\subsubsection{Session 1}

First, participants read and signed a consent form. Then, the experimenter used a temperature probe to record their right index finger temperature. Temperature was recorded every 2 minutes across a 10 minute period, for a total of 6 measurements. These were averaged to calculate the required cold wood temperature. Participants were not told why their finger temperature was taken. The cold wood materials were prepared accordingly for Session 2, which was scheduled on a different day to allow sufficient cooling time.

\subsubsection{Session 2}

An experimenter followed a script to explain the experimental protocol and show how to touch the materials. Ambient temperature wood and aluminum materials were used in the demonstration and practice.  Fig. \ref{fig:3} shows the human protocol for a single experiment trial. Participants were told that they would:

\begin{itemize}
    \item touch metal and wood materials and be asked to identify them
    \item have 3 experiment trials with 3 material touching cases per trial
    \item don a blindfold -- Fig. \ref{fig:3}(a) 
    \item don noise-cancelling headphones -- Fig. \ref{fig:3}(b), that would play back the prompts shown in Fig. \ref{fig:3}(f),(j),(l), and (m), in addition to white noise
    \item use their right index finger to touch the materials -- Fig. \ref{fig:3}(c)
    \item have their finger temperature taken before and after every trial
\end{itemize}

A $2''\times2 ''$ black square of foam padding was adhered to the table underneath the desired elbow position to help the participants prepare their arm while blindfolded, as shown in fig \ref{fig:3}(d). The researcher recited the prompts aloud while demonstrating how to touch the materials. The researcher provided the following instructions:
\begin{itemize}
    \item When preparing to touch a material, keep the right forearm resting on the table.
    \item Keep the left arm out of the way so it does not interfere -- Fig. \ref{fig:3}(e).
    \item Upon hearing `In 5 seconds I will ask you to touch the material,' bend hand up at the wrist and prepare the finger immediately -- Fig. \ref{fig:3}(f).
    \item Make a tight fist with the three rightmost fingers and the thumb -- Fig. \ref{fig:3}(g).
    \item Curve the index finger into a slight arch held 3-5 inches above the table -- Fig. \ref{fig:3}(h).
    \item When `Touch' is heard, draw the index finger down steadily to touch the material -- Fig. \ref{fig:3}(j),(k).
    \item Do not move the finger across the surface of the material.
    \item Do not move the finger too quickly or push the material.
    \item Take the meaning of touch at face value.
    \item When `Release' is heard, draw the hand back up and answer the following question -- Fig. \ref{fig:3}(l),(m).
    \item Wait to remove the blindfold and headphones until the researcher turns off the white noise following the 3rd and final touch in a trial -- Fig. \ref{fig:3}(p).
    \item Between trials, there is a 5 minute rest period where talking and questions are discouraged.
    \item The experiment is not a game.
\end{itemize}


Within trials, the $\sim 1$ minute pause between material touches gave the experimenter time to prepare the next material. Between trials, participants removed the blindfold and headphones for a $\sim 5$ minute rest period to (1) increase independence between trials and (2) alleviate potential dizziness or disorientation from having their senses blocked for too long.


After performing the demonstration, the researcher got out of the chair in which participants sit during the study and switched places with the participant. The participant then began a brief period of practice with ambient temperature wood and metal to ensure adherence to the protocol. This also gave participants experience touching the material samples at the ambient temperature. Practice included at least 3 material touches where the researcher read the prompts aloud, giving sufficient time to correct the participant if instructions were followed improperly or forgotten. Then, the participant donned the headphones and performed at least two practice trials with ambient temperature metal and wood without a blindfold. Some participants required additional practice touches to either feel comfortable or ensure they were touching the material according to the instructions. 

After the participant practiced, the actual trials began. 

\begin{figure*}[!t]
\begin{center}
\includegraphics[width=18cm]{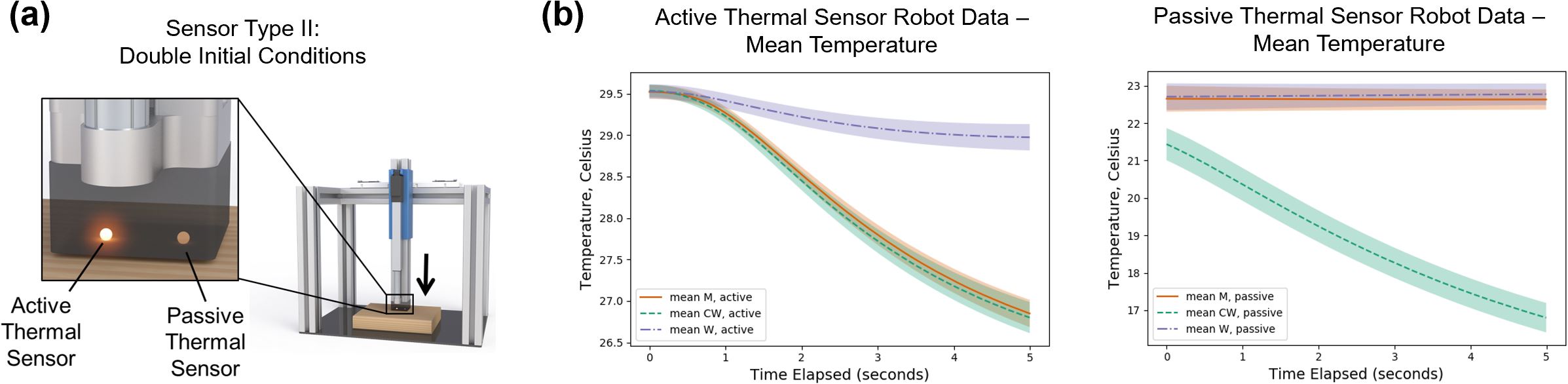} 
\vspace{-2mm}
\caption{\label{fig:4}\textbf{Data collected by the robot.} (a) We equipped the robot with double initial condition sensing to be able to outperform humans, i.e. the robot has two sensors initially at different temperatures. (b) Data collected by the active (heated) sensor shows a high degree of similarity between cold wood and ambient temperature metal (left). Data collected by the passive (unheated) sensor shows a high degree of similarity between ambient temperature materials, but the cold wood data is substantially different.
}
\end{center}
\end{figure*} 


\subsection{Similarity Test}

We conducted a human study to test similarity between materials at ambiguous conditions. Historically, two inference approaches have been used for showing similarity of conditions: the power approach and variants of the two one-sided test.  However, the power approach was found to be inappropriate for similarity testing \cite{schuirmann1987comparison,fda1992us}.  The two one-sided test has been extended to similarity of two proportions, first introduced by Dunnett and Gent \cite{dunnett1977significance}. For this test, some similarity margin is chosen and inference is made based on the probability of both samples falling within this margin.  However, a matched pair design lacks independence. For difference testing, McNemar \cite{mcnemar1947note} proposed a term to correct for matched pair dependence and Eliasziw and Donner \cite{eliasziw1991application} followed with a correction for clustered-paired data. For similarity testing, Nam and Kwon \cite{nam2009non} proposed a modification of Obuchowski's method of clustered-paired similarity testing \cite{obuchowski1998on}, and showed that this performs well in studies with a cluster number less than 100, where the number of pairs in a cluster is less than five. In a more recent work \cite{nam2011power}, they formulate power and sample size requirements that we employ. 

Using the method of \cite{nam2011power}, for a similarity margin of $\delta =0.2$, we require 29 participants (rounded to 30) assuming that 2 out of 3 clustered paired trials will feature a finger temperature within the range $\mathcal{T}_\mathrm{f} \pm \gamma$.  We require that each participant in $N = 30$ have at least 1 out of 3 trials with their finger temperature in range; otherwise N is increased. Two of the participants did not have any finger temperatures within $\mathcal{T}_\mathrm{f} \pm \gamma$, so these were excluded from inference, leading to a total of $\tilde{N}=32$. Of the 30 participants with at least one trial in range, the average number of trials in range was 2.67.




\subsection{Handling Temperature Variations}

We found that the cold wood warmed substantially in the time after it was taken out of the refrigerator before the participant touched it. A correction factor was used to account for warming that occurred between removing the cold wood from the refrigerator and placing it before the participant to touch. To determine refrigerator temperature, we fit a linear relationship between the initial temperature at $t=0s$ and touched temperature at $t=20s$ that depends on the nominal value of $\mathcal{T}_\mathrm{fridge}$, by recording each at $5^\circ$ intervals between $-35^\circ C$ and $20^\circ C$. 

In a pilot study, we found that finger temperature can vary substantially between trials. Thus, we monitored finger temperature between trials and withdrew from statistical inference the trials where the finger temperature deviates too far from the ambiguous condition. Finger temperature was recorded just prior to and following a trial; we computed the average of these to represent each trial. Participants were not shown or told their finger temperature. To calculate $\gamma$, the deviation threshold, we used data from 13 pilot participants and cooled the wood to $7^\circ C$, which is thermally ambiguous when the finger is $\widehat{\mathcal{T}_\mathrm{f}} = 27^\circ C$. We chose $\widehat{\mathcal{T}_\mathrm{f}}$ from the mean human finger temperature measured in~\cite{montgomery1976effect}. To find the allowable threshold, we maximized the proportion $p_{mm} | \mathcal{T}_\mathrm{f}' \in \beta$  of answering ‘metal’ for both cold wood and metal given that the finger temperature is within the range and, and the proportion  $p_{w} | \mathcal{T}_\mathrm{f}' \not\in \beta$ of answering ‘wood’ for at least one of these two paired samples given that the finger temperature is outside the range:

\begin{equation*}
\operatornamewithlimits{argmax}\limits_{p_{mm|\mathcal{T}_\mathrm{f}'\in \beta} + p_{w|\mathcal{T}_\mathrm{f}'\not\in \beta} } f\Big(\beta=\Big[\widehat{\mathcal{T}_\mathrm{f}} - \gamma, \widehat{\mathcal{T}_\mathrm{f}} + \gamma \Big] \text{ s.t.  } \gamma = {\scriptstyle 0.0,0.5,1.0,...}\Big)
\end{equation*}

We found that $\gamma = 3.5^{\circ}C$.  

\section{Three Robot Studies} 

We conducted three robot studies. The first robot study models the human participants in our human study. The second robot study investigates how well robots, and possibly people, can perform given ambiguous conditions with which they have received specific training. The third robot study empirically evaluates the ability of a double-condition sensor to overcome ambiguous conditions. 

\subsection{Study Design}

\subsubsection{Robot Study 1: A Human-like Sensor and Training with Ambient Temperature Materials} 

For the first robotic investigation, we conducted a study in which we equipped a robot with a human-like sensor (Fig.~\ref{fig:2} (b)), and gave the robot the same material recognition problem as the human participants. %
We set the initial temperature, $\mathcal{T}_{\mathrm{r}}$, for the robot's actively heated sensor to be the mean of the human finger temperatures we measured in the first human study sessions. When we apply our thermal sensing model to robotic perception, the sensor temperature, $\mathcal{T}_{\mathrm{s}}$, is equal to the robot's sensor temperature, $\mathcal{T}_{\mathrm{s}}=\mathcal{T}_{\mathrm{r}}$. The robot contacted the materials with a target contact force of 3N. Once in contact, the temperature of the robot's sensor decreased as heat transferred from it into the material. When working with temperature time series from the robot's sensor, we estimated the time of contact to occur at the temperature peak just as the heat began transferring into the material. 



\subsubsection{Robot Study 2: A Human-like Sensor and Training with Ambient Temperature and Cooled Materials}

For the second robot study, we used the same protocol as in Robot Study 1, but trained the robot's material classifier using data from all material conditions: cold wood, ambient wood, and ambient metal. Data-driven models have the potential to find informative differences between the ambiguous conditions (i.e. the cold wood and metal curve in Fig. \ref{fig:4}(b)-left) that our idealized mathematical model neglects. 



\subsubsection{Robot Study 3: A Double-Condition Sensor and Training with Ambient Temperature and Cooled Materials} 

For the third robot study, we created a double-condition sensor that our mathematical model predicted would overcome ambiguous conditions. The specific design we evaluated uses two adjacent thermal sensors, one actively heated and one passive (unheated), which are at two different initial conditions. Fig. \ref{fig:4}(a) shows the sensor module with the two adjacent sensors. 



For the second sensor initial conditions, we used a passive unheated sensor to reduce the complexity of our design. Fig.~\ref{fig:4}(b)-right shows data collected by the passive thermal sensor using the double-condition sensor. With this data, as well as the active sensor data, a robot can overcome the ambiguity. We trained a data-driven model on both active and passive sensor data for all conditions: cold wood, ambient wood, and ambient metal.



\subsection{Data Collection}

The three robot studies use data for training and testing acquired during the same session. The first two robot studies use temperature measurements from a single heated temperature sensor. The third robot study uses temperature measurements from the same heated temperature sensor and an additional unheated temperature sensor. Using data from the same session makes the results from the three robot studies more directly comparable by eliminating variations that could occur between sessions, such as due to sensor and actuator noise. 

We collected data for the three robot studies by having a robot perform $N=30$ sets of three trials in a manner consistent the human study. As with the human study, each trial consisted of contact with three material samples: ambient temperature metal, ambient temperature wood, and cold wood. Fig. \ref{fig:4}-(a) shows the 1-DoF robot used to collect temperature measurements from both a heated and an unheated sensor. As such, the robot collected $540 ~\mathrm{time~series}= 30 ~\mathrm{sets} \times 3 ~\mathrm{trials} \times 3 ~\mathrm{material~samples} \times 2 ~\mathrm{temperature~sensors}$. Prior to contact, we heated the robot's active sensor to the mean human finger temperature measured in the first human study session ($\overline{\mathcal{T}_\mathrm{f}} = 29.5^\circ C$). A description and schematic of the robot electronic components can be found in Appendix B. 



\begin{figure}
\begin{center}
\includegraphics[width=6.0cm]{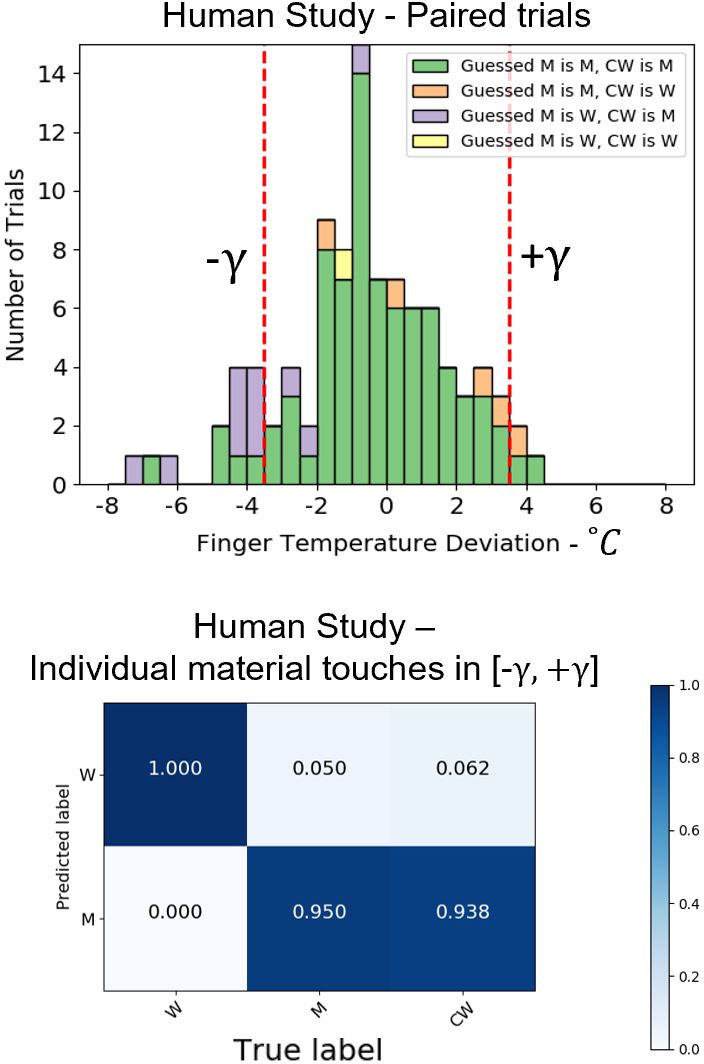} 
\vspace{-2mm}
\caption{\label{fig:5}\textbf{Results - Human Study.} Humans consistently misclassified cold wood as metal in our study of materials at ambiguous conditions when their finger temperature was near ambiguous conditions ($\mathcal{T}_\mathrm{f} \pm \gamma$), with high significance of similarity between aluminum and cold wood. }
\end{center}
\end{figure}

\subsection{Creating Ambiguous Conditions for the Robot}



To create ambiguous conditions for the robot, we first used the mathematical model to estimate the ambiguous temperature of cold wood, $\mathcal{T}_\mathrm{cw}$. To account for temperature variation due to experimental delays and other factors, we performed an empirical search for a nearby value of $\mathcal{T}_\mathrm{cw}$ that would result in high ambiguity. Specifically, we recorded temperatures over time resulting from the robot touching wood cooled by adjusting the refrigerator temperature by $1^\circ C$ increments around the mathematically calculated value of $\mathcal{T}_\mathrm{cw}$. We collected many samples and computed the mean temperature curves over time. We then selected the refrigerator temperature that resulted in the most visually similar mean curves by plotting mean curves from cold wood and comparing them with a plotted mean curve from ambient temperature metal (see Fig.~\ref{fig:4}(b)-left). The mathematically predicted temperature target was $5.29^\circ C$, while the refrigeration target temperature we selected using this process was $6^\circ C$. 


Unlike the human study that required $t\approx20 s$ between refrigerator removal and human touch, we placed the robot adjacent to the refrigerator to decrease the delay to $t\approx5 s$ between controlled refrigeration and the robot making contact with the material. Due to this methodology and the greater control afforded by the use of a robot, we expect that the conditions for the robot studies were more ambiguous than the conditions for the human study.

\begin{figure}
\begin{center}
\includegraphics[width=8.5cm]{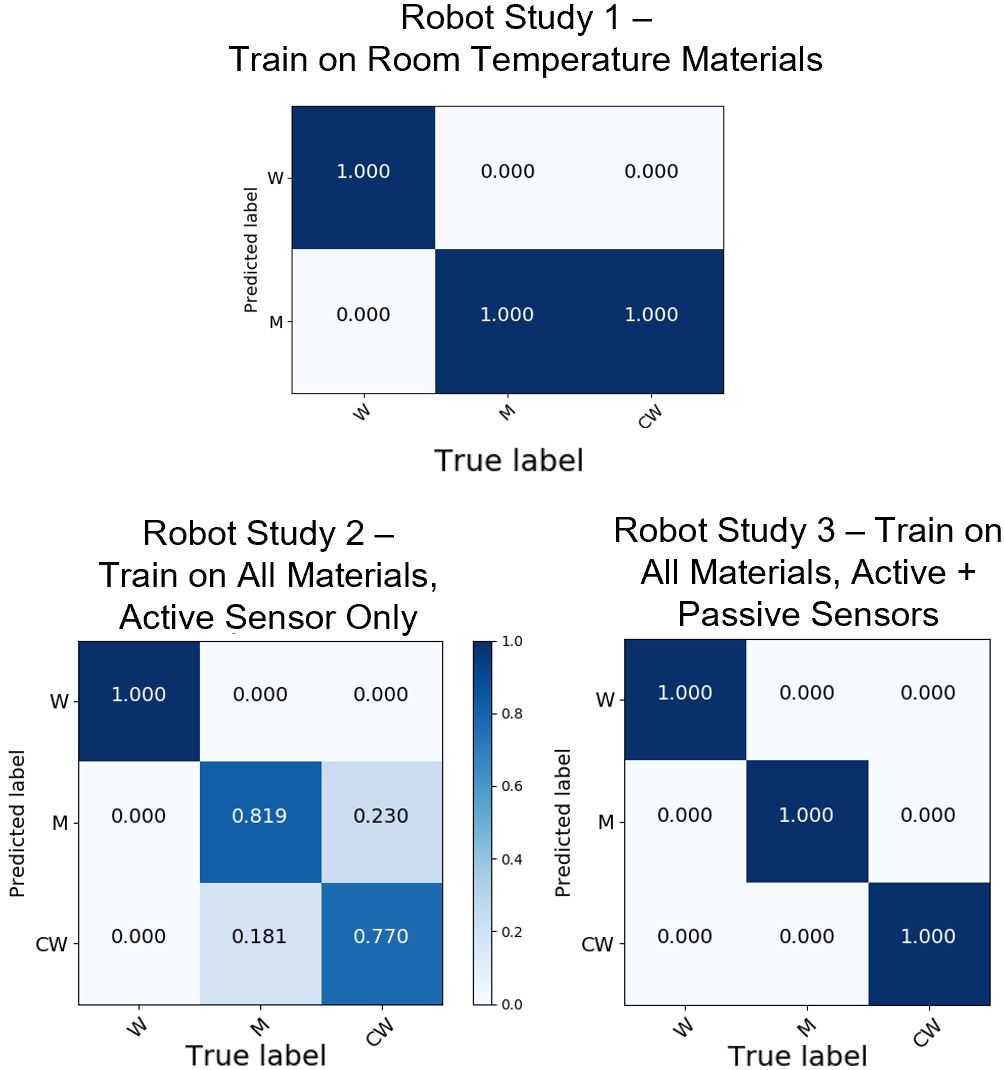} 
\vspace{-2mm}
\caption{\label{fig:6}\textbf{Results - Robot Study.} Robots equipped with a human-like sensor and trained on ambient temperature materials made the same error as humans (top). Using a human-like sensor, the robot was sometimes able to correctly classify thermally ambiguous cold wood and metal when trained on all material conditions (bottom left). Using the double-condition sensor entirely eliminates ambiguity (bottom right). }
\end{center}
\end{figure}

\section{Results}

We report results from each of the studies below.

\subsection{Results from the Human Study} 

\subsubsection{Humans confuse thermally ambiguous materials}

We used a clustered-paired similarity test to show similarity of conditions \cite{nam2009non}. Our results from 32 participants show high statistical significance that wood feels similar to metal when wood is cooled to a thermally ambiguous temperature, $\mathcal{T}_\mathrm{o2} \approx \mathcal{T}_\mathrm{cw}$. $\mathcal{T}_\mathrm{cw}$ is based on the finger temperature measured in the first session, $\mathcal{T}_\mathrm{s}=\mathcal{T}_\mathrm{f}$, and calculated with Eq.~\ref{eq:4}. For each trial, we measured finger temperature $\mathcal{T}_\mathrm{f}'$, where $\epsilon = \mathcal{T}_\mathrm{f} - \mathcal{T}_\mathrm{f}'$ is the deviation from the intended sensing condition. The histogram in Fig. \ref{fig:5} shows this with $-\gamma \leq \epsilon \leq +\gamma$ bounds to determine if $\mathcal{T}_\mathrm{f}'$ is sufficiently close to $\mathcal{T}_\mathrm{f}$, where $\gamma$ is the maximum allowable deviation. We find $\gamma = 3.5^\circ C$ using optimization.

\subsubsection{Deviation from ambiguous conditions affects human confusion}

The types of errors that participants made were associated with differences between participants' actual finger temperatures, $\mathcal{T}_\mathrm{f}'$, and the expected finger temperatures, $\mathcal{T}_\mathrm{f}$, used to calculate the intended ambiguous conditions (i.e. the cold wood temperatures, $\mathcal{T}_\mathrm{cw}$). We found that when a person's finger temperature was greater than $\mathcal{T}_\mathrm{f}$ they were more likely to guess that the cold wood was wood and the metal was metal. When the finger was less than $\mathcal{T}_\mathrm{f}$, they were more likely to guess that the cold wood was metal and the metal was wood.

\subsection{Results from the Robot Studies} 

\subsubsection{Robot Study 1: When trained only on ambient conditions, robots confuse cold wood and metal}

We trained a linear kernel SVM classifier on robot data for ambient temperature materials, and tested the classifier on these materials in addition to cold wood, shown in Fig.~\ref{fig:6}-top. For each set of clustered paired trials, there were a total of 9 unique material blocks: Each material type was always tested in the same trial order, i.e., for trial 1, the same wood ($W1$), cold wood ($CW1$), and metal ($M1$) blocks were always used in whatever random order was chosen. We performed 27-fold leave-one-\textit{block}-out cross validation wherein each fold featured training on all data from two unique blocks of both ambient temperature materials (i.e. $30 \times M1$, $30 \times M2$, $30 \times W1$, $30 \times W2$) and testing on the remaining block of all materials (i.e. $30 \times M3$, $30 \times W3$, and $30 \times CW1$). With this training, the robot consistently classified cold wood as metal.


\subsubsection{Robot Study 2: Training the robot on multiple temperatures can reduce ambiguity}

\begin{figure}
\begin{center}
\includegraphics[width=9.3cm]{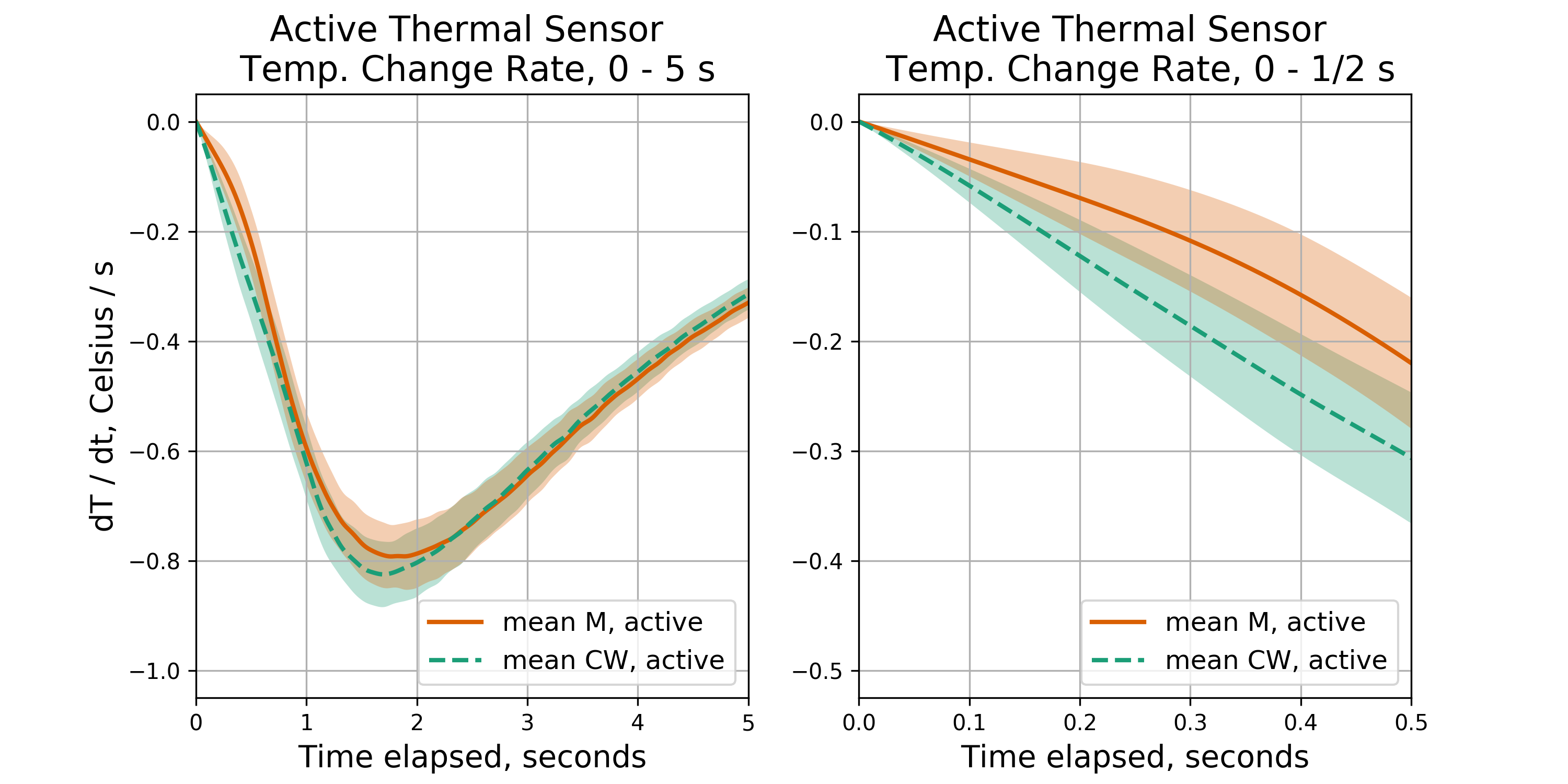} 
\vspace{-4mm}
\caption{\label{fig:7}\textbf{Subtle discrepancies in thermally ambiguous materials.} Differences in the distributions of the slopes may account for classification success when a robot with human-like sensing capabilities is trained on all material conditions (Recall Fig.~\ref{fig:6}-bottom left). (left) Mean and standard deviation for signal slopes over the full time range used for classification. (right) Early in the process, the slope distributions differ conspicuously.}
\end{center}
\end{figure}

Fig. \ref{fig:6}(b)-lower left shows results for robot with a human-like sensor from 27-fold leave-one-\textit{block}-out cross validation with inclusion of the cold wood blocks during training of linear kernel SVMs. This additional training data resulted in significantly better classification performance. We conducted a post hoc analysis of the data in an attempt to find noticeable differences between the ambiguous conditions that might be used to correctly classify the materials. We identified visually apparent differences in the slopes of the temperature measurements over time from the ambiguous conditions, as shown in Fig. \ref{fig:7}. The discrepancy is greatest during the first 1/2 second of contact, where contact is assumed to occur at the maximum point of the temperature curve. Our findings showing that humans are unable to exploit these phenomena is consistent with past research showing humans' inability to detect small changes in temperature \cite{stevens1998temperature}. 



\subsubsection{Robot Study 3: Using the double-condition sensor resolves the ambiguity}

Using an active and a passive sensor, the robot classified the materials with $100\%$ accuracy as shown in Fig. \ref{fig:6}-bottom right, using a linear kernel SVM classifier.

\section{Discussion}


Our results provide insights into human and robot perception of materials via heat transfer given ambiguous conditions. 

\subsection{Human Errors from Ambiguous Conditions}

Our results with human participants confirms the existence of ambiguous conditions when humans touch common materials, and provides evidence that the simplified heat transfer model can predict conditions that confuse people. 

The mean human finger temperature measured during the material recognition trials was lower ($\overline{\mathcal{T}_\mathrm{f}'} = 28.5 ^\circ C$) than that measured in the first session ($\overline{\mathcal{T}_\mathrm{f}} = 29.5^\circ C$). This explains why there are more trials left of center in the histogram in Fig. \ref{fig:5} -- i.e., $\epsilon < 0$ in 60 trials, $\epsilon > 0$ in 32 trials. This is possibly due to finger temperature slightly dropping during the experiment because of extended open-hand air exposure during the pre-experiment prep, whereas their finger temperature was measured sooner during the first session (e.g. after they had hands in pockets or using a phone). 

\subsection{Robot Errors from Ambiguous Conditions}

Our first robot study was analogous to the human study. We trained the robot with data from contact with ambient temperature materials, which is similar to how human participants practiced with ambient temperature materials prior to the experiment. The first robot study demonstrates that ambiguous conditions can also confuse robots, even with carefully controlled situations for which they have been trained. It also demonstrates that the simplified heat transfer model can be used to predict situations that will confuse robots. 

Our second robot study indicates that robots, and possibly humans, can partly resolve ambiguities when trained on ambiguous examples. The robot likely used subtle differences to correctly classify the ambiguous materials that are not represented by the simplified heat transfer model. The physical phenomena responsible for these differences remain unclear. Our post hoc analysis suggests that differences at the beginning of contact may have played a role. Heat transfer prior to contact due to convection while the robot's finger moved slowly into contact may be a factor. Notably, in the human studies people moved their fingers more quickly into contact. Identifying the factors at work might provide new avenues for improving robotic performance. 

While we trained robots with ambiguous examples, we did not do something similar for humans. Like robots, it is plausible that people might be able to learn to overcome the ambiguity by taking advantage of subtle cues. Whether human perception is capable of making such distinctions remains an open question. 

\subsection{Naturalistic Sensing}

Under more naturalistic and diverse conditions, we would expect the impact of ambiguous conditions to be more severe. We only considered a controlled material recognition problem with carefully prepared samples representing two materials. Performance would likely diminish under real-world conditions with more material types, temperature variations, and other forms of variability. To some extent, successful recognition in the second robot study may relate to overfitting to the data from our controlled study. 

\subsection{Robots Outperforming Humans}

We would expect for the double-condition sensor used in the third robot study to perform well under real-world conditions by directly overcoming the ambiguity predicted by the simplified heat transfer model. Our mathematical proof and our third robot study provide strong evidence that the double-condition sensor overcomes ambiguous initial conditions. 

In prior work, Kerr \textit{et al.}~\cite{kerr2013material} found that robots can outperform human participants when using heat transfer to recognize materials at ambient temperatures. Robots have also outperformed humans at texture discrimination using exploratory movements with non-thermal modalities \cite{fishel2012bayesian}. Our research suggests that with double-condition sensors robots can outperform human perception via heat transfer given ambiguous initial conditions. Together, these results suggest that robots have the potential to outperform human touch sensing across diverse haptic modalities and conditions.

\section*{Acknowledgment}

We thank Linda Komnang Liezu for her work in developing the robot sensor. 

This work was supported by the National Science Foundation (NSF) Emerging Frontiers in Research and Innovation (EFRI) award 1137229, NSF award IIS-1150157, NSF Graduate Research Fellowship Program under Grant No. DGE-1148903, and NSF SURE Robotics REU award number EEC-1263049.

\ifCLASSOPTIONcaptionsoff
  \newpage
\fi



\bibliographystyle{IEEEtran}
%

\bibliography{thermal}

\begin{IEEEbiography}[{\includegraphics[width=1in,height=1.25in,clip, keepaspectratio]{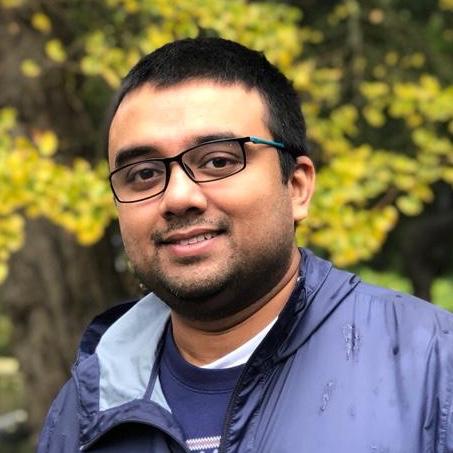}}]{Tapomayukh Bhattacharjee} is an NIH Ruth L. Kirschstein NRSA postdoctoral research associate in Computer Science and Engineering at the University of Washington. He completed his Ph.D. in Robotics from Georgia Institute of Technology, received his M.S. from Korea Advanced Institute of Science and Technology (KAIST), Daejeon, South Korea, and B.Tech. from National Institute of Technology, Calicut, India. His primary research interests are in the fields of human-robot interaction, haptic perception, and robot manipulation. His research revolves around the theme of leveraging physical interactions with objects and humans in unstructured environments to enable assistive care for people with mobility limitations.
\end{IEEEbiography}

\begin{IEEEbiography}[{\includegraphics[width=1in,height=1.25in,clip, keepaspectratio]{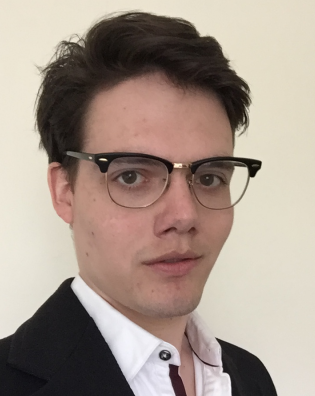}}]{Henry M. Clever}
received a B.S. in mechanical engineering from the University of Kansas in 2014, and a M.S. in mechanical engineering from New York University in 2016. Henry is currently pursuing a Ph.D. in robotics at the Georgia Institute of Technology in the Healthcare Robotics Lab. 

Henry's reseach interests include robot understanding in unstructured environments, haptic and vision perception of humans and robots, human-robot systems, physics simulation of humans and robots and human pose estimation. Henry's awards include a NSF G-K12 Teaching Fellowship (2014), an NSF Graduate Research Fellowship (2015), a Georgia Tech President's Fellowship (2016), and a NSF Research Traineeship in human-centered robotics (2016).
\end{IEEEbiography}

\begin{IEEEbiography}[{\includegraphics[width=1in,height=1.25in,clip, keepaspectratio]{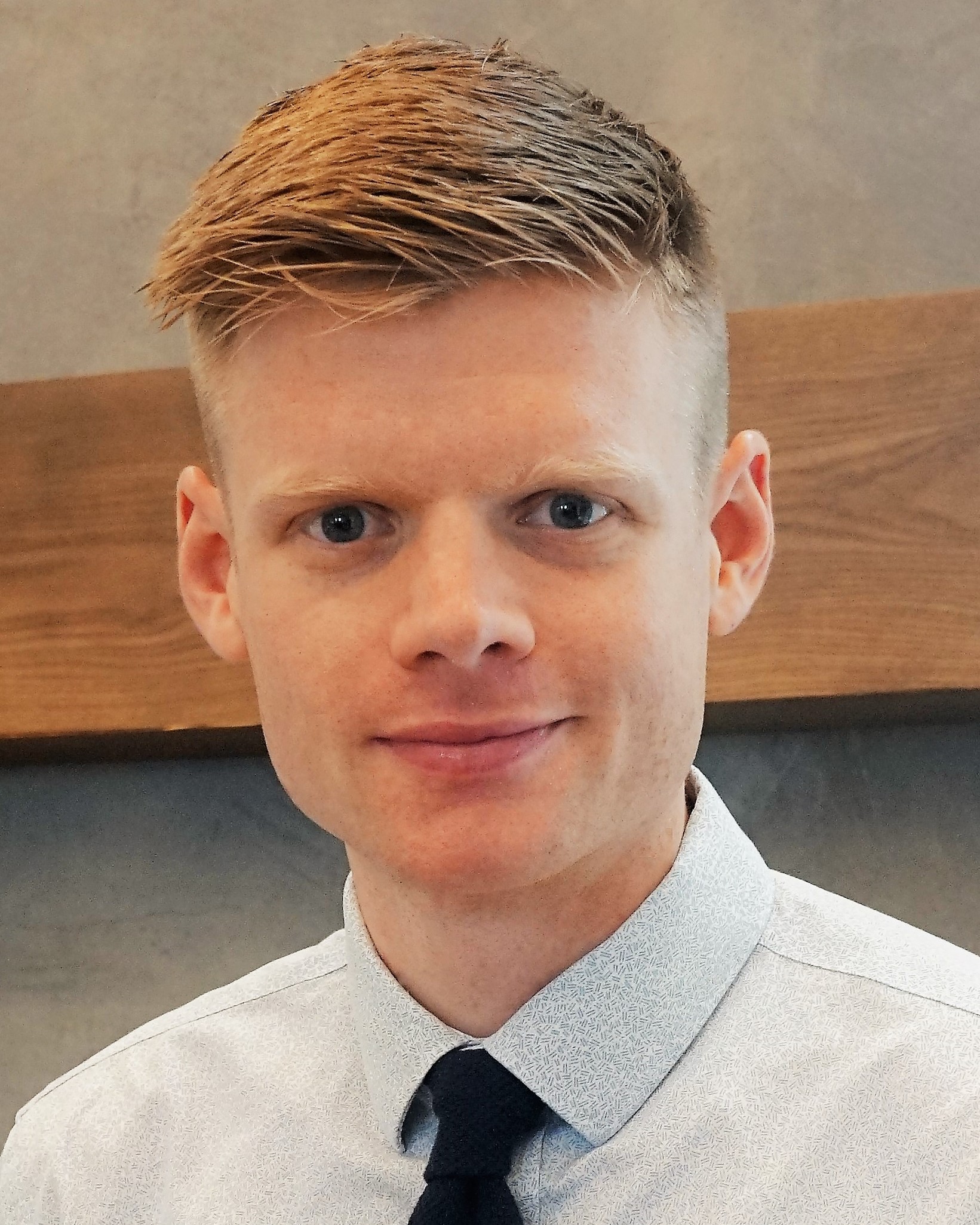}}]{Joshua Wade} is a systems engineer for the U.S. Navy. He completed a M.S. in mechanical engineering at the Georgia Institute of Technology where he conducted research in the Healthcare Robotics Lab. Josh graduated with a B.S. in mechanical engineering from the Georgia Institute of Technology where he recivied the 2016 Tau Beta Pi Outstanding Graduate award. His research interests include robotics, tactile sensing, mechatronics, and controls.
 
\end{IEEEbiography}

\begin{IEEEbiography}[{\includegraphics[width=1in,keepaspectratio]{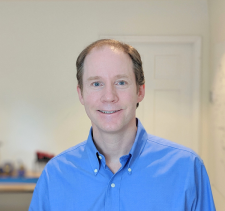}}]{Charles C. Kemp} (Charlie Kemp) is an Associate Professor at Georgia Tech in the Department of Biomedical Engineering with adjunct appointments in the School of Interactive Computing and the School of Electrical and Computer Engineering. In 2007, he founded the Healthcare Robotics Lab, which focuses on enabling robots to provide intelligent physical assistance in the context of healthcare. He earned a BS, an MEng, and a PhD from the Massachusetts Institute of Technology (MIT) in the areas of computer science and electrical engineering.  


\end{IEEEbiography}





\end{document}


\newcommand{\red}{\color[rgb]{0.6,0.0,0.0}}
\newcommand{\yellow}{\color[rgb]{0.6,0.6,0.0}}
\newcommand{\green}{\color[rgb]{0.0,0.5,0.0}}
\newcommand{\blue}{\color[rgb]{0.0,0.4,0.6}}
\newcommand{\purple}{\color[rgb]{0.6,0.15,0.7}}

\newcommand{\josh}[1]{{[\bf \blue {Josh: } #1}]}
\newcommand{\hc}[1]{{[\bf \green {Henry C: } #1}]}
\newcommand{\ck}[1]{{[\bf \red {Charlie: } #1}]}
\newcommand{\tapo}[1]{{[\bf \purple {Tapo: } #1}]}

%

\title{\huge{Material Recognition via Heat Transfer Given \\ Ambiguous Initial Conditions - ***Appendix***}}




%
%
%

 \author{Tapomayukh Bhattacharjee*,~\IEEEmembership{Student Member,~IEEE,} Henry M.~Clever*$^{\dagger}$,~\IEEEmembership{Student Member,~IEEE,} Joshua~Wade,~\IEEEmembership{Student Member,~IEEE,} Charles C.~Kemp,~\IEEEmembership{Member,~IEEE}
\thanks{All authors were with the Department
of Department of Biomedical Engineering, Georgia Institute of Technology, Atlanta,
GA, 30332 USA.}
\thanks{*Authors contributed equally.} \thanks{$^{\dagger}$Corresponding author, email: henryclever@gatech.edu.  }
\thanks{Manuscript received October XX 2020; revised XXXXX.}}

\markboth{Journal of \LaTeX\ Class Files,~Vol.~14, No.~8, August~2015}%
{Shell \MakeLowercase{\textit{et al.}}: Bare Demo of IEEEtran.cls for IEEE Journals}
%

\maketitle
\setcounter{figure}{7}

\IEEEpeerreviewmaketitle

\appendices
\section{Proof for overcoming ambiguity.}
Here we present the proof for the inequality presented in Eq.~5. Starting from Eq.~4, with two sensors with initial temperatures $\mathcal{T}_{\mathrm{s1}}$ and $\mathcal{T}_{\mathrm{s2}}$ and effusivities $e_{\mathrm{s1}}$ and $e_{\mathrm{s2}}$,

\vspace{4mm}

\setlength{\parindent}{0cm}
\large{${-\mathcal{T}_{\mathrm{s1}}e_{\mathrm{s1}}e_{\mathrm{o1}} + \mathcal{T}_{\mathrm{s1}}e_{\mathrm{s1}}e_{\mathrm{o2}} + \mathcal{T}_{\mathrm{o1}}e_{\mathrm{s1}}e_{\mathrm{o1}} + \mathcal{T}_{\mathrm{o1}}e_{\mathrm{o1}}e_{\mathrm{o2}} \over e_{\mathrm{o2}}(e_{\mathrm{s1}} + e_{\mathrm{o1}})} \neq {-\mathcal{T}_{\mathrm{s2}}e_{\mathrm{s2}}e_{\mathrm{o1}} + \mathcal{T}_{\mathrm{s2}}e_{\mathrm{s2}}e_{\mathrm{o2}} + \mathcal{T}_{\mathrm{o1}}e_{\mathrm{s2}}e_{\mathrm{o1}} + \mathcal{T}_{\mathrm{o1}}e_{\mathrm{o1}}e_{\mathrm{o2}} \over e_{\mathrm{o2}}(e_{\mathrm{s2}} + e_{\mathrm{o1}})} $}

\vspace{4mm}

\small{$(-\mathcal{T}_{\mathrm{s1}}e_{\mathrm{s1}}e_{\mathrm{o1}} + \mathcal{T}_{\mathrm{s1}}e_{\mathrm{s1}}e_{\mathrm{o2}} + \mathcal{T}_{\mathrm{o1}}e_{\mathrm{s1}}e_{\mathrm{o1}} + \mathcal{T}_{\mathrm{o1}}e_{\mathrm{o1}}e_{\mathrm{o2}})(e_{\mathrm{s2}} + e_{\mathrm{o1}}) \neq$

$(-\mathcal{T}_{\mathrm{s2}}e_{\mathrm{s2}}e_{\mathrm{o1}} + \mathcal{T}_{\mathrm{s2}}e_{\mathrm{s2}}e_{\mathrm{o2}} + \mathcal{T}_{\mathrm{o1}}e_{\mathrm{s2}}e_{\mathrm{o1}} + \mathcal{T}_{\mathrm{o1}}e_{\mathrm{o1}}e_{\mathrm{o2}})(e_{\mathrm{s1}} + e_{\mathrm{o1}}) $

\vspace{4mm}
$-\mathcal{T}_{\mathrm{s1}}e_{\mathrm{s1}}e_{\mathrm{s2}}e_{\mathrm{o1}} + \mathcal{T}_{\mathrm{s1}}e_{\mathrm{s1}}e_{\mathrm{s2}}e_{\mathrm{o2}} + \mathcal{T}_{\mathrm{o1}}e_{\mathrm{s1}}e_{\mathrm{s2}}e_{\mathrm{o1}} + \mathcal{T}_{\mathrm{o1}}e_{\mathrm{s2}}e_{\mathrm{o1}}e_{\mathrm{o2}} -\mathcal{T}_{\mathrm{s1}}e_{\mathrm{s1}}e_{\mathrm{o1}}^2 + \mathcal{T}_{\mathrm{s1}}e_{\mathrm{s1}}e_{\mathrm{o1}}e_{\mathrm{o2}} + \mathcal{T}_{\mathrm{o1}}e_{\mathrm{s1}}e_{\mathrm{o1}}^2 + \mathcal{T}_{\mathrm{o1}}e_{\mathrm{o1}}^2e_{\mathrm{o2}}  \neq  -\mathcal{T}_{\mathrm{s2}}e_{\mathrm{s1}}e_{\mathrm{s2}}e_{\mathrm{o1}} + \mathcal{T}_{\mathrm{s2}}e_{\mathrm{s1}}e_{\mathrm{s2}}e_{\mathrm{o2}} + \mathcal{T}_{\mathrm{o1}}e_{\mathrm{s1}}e_{\mathrm{s2}}e_{\mathrm{o1}} + \mathcal{T}_{\mathrm{o1}}e_{\mathrm{s1}}e_{\mathrm{o1}}e_{\mathrm{o2}} 
-\mathcal{T}_{\mathrm{s2}}e_{\mathrm{s2}}e_{\mathrm{o1}}^2 + \mathcal{T}_{\mathrm{s2}}e_{\mathrm{s2}}e_{\mathrm{o1}}e_{\mathrm{o2}} + \mathcal{T}_{\mathrm{o1}}e_{\mathrm{s2}}e_{\mathrm{o1}}^2 + \mathcal{T}_{\mathrm{o1}}e_{\mathrm{o1}}^2e_{\mathrm{o2}} $

\vspace{4mm}
$-\mathcal{T}_{\mathrm{s1}}e_{\mathrm{s1}}e_{\mathrm{s2}}e_{\mathrm{o1}} + \mathcal{T}_{\mathrm{s1}}e_{\mathrm{s1}}e_{\mathrm{s2}}e_{\mathrm{o2}} + \mathcal{T}_{\mathrm{o1}}e_{\mathrm{s2}}e_{\mathrm{o1}}e_{\mathrm{o2}} -\mathcal{T}_{\mathrm{s1}}e_{\mathrm{s1}}e_{\mathrm{o1}}^2 + \mathcal{T}_{\mathrm{s1}}e_{\mathrm{s1}}e_{\mathrm{o1}}e_{\mathrm{o2}} + \mathcal{T}_{\mathrm{o1}}e_{\mathrm{s1}}e_{\mathrm{o1}}^2 \neq$

$ -\mathcal{T}_{\mathrm{s2}}e_{\mathrm{s1}}e_{\mathrm{s2}}e_{\mathrm{o1}} + \mathcal{T}_{\mathrm{s2}}e_{\mathrm{s1}}e_{\mathrm{s2}}e_{\mathrm{o2}} + \mathcal{T}_{\mathrm{o1}}e_{\mathrm{s1}}e_{\mathrm{o1}}e_{\mathrm{o2}} 
-\mathcal{T}_{\mathrm{s2}}e_{\mathrm{s2}}e_{\mathrm{o1}}^2 + \mathcal{T}_{\mathrm{s2}}e_{\mathrm{s2}}e_{\mathrm{o1}}e_{\mathrm{o2}} + \mathcal{T}_{\mathrm{o1}}e_{\mathrm{s2}}e_{\mathrm{o1}}^2 \ $

\vspace{4mm}
$\mathcal{T}_{\mathrm{s1}}\big(-e_{\mathrm{s1}}e_{\mathrm{s2}}e_{\mathrm{o1}} + e_{\mathrm{s1}}e_{\mathrm{s2}}e_{\mathrm{o2}} - e_{\mathrm{s1}}e_{\mathrm{o1}}^2 + e_{\mathrm{s1}}e_{\mathrm{o1}}e_{\mathrm{o2}}\big) + \mathcal{T}_{\mathrm{o1}}\big(e_{\mathrm{s2}}e_{\mathrm{o1}}e_{\mathrm{o2}}  + e_{\mathrm{s1}}e_{\mathrm{o1}}^2 \big) \neq$

$  \mathcal{T}_{\mathrm{s2}}\big(-e_{\mathrm{s1}}e_{\mathrm{s2}}e_{\mathrm{o1}} + e_{\mathrm{s1}}e_{\mathrm{s2}}e_{\mathrm{o2}} -e_{\mathrm{s2}}e_{\mathrm{o1}}^2 + e_{\mathrm{s2}}e_{\mathrm{o1}}e_{\mathrm{o2}}\big) +  \mathcal{T}_{\mathrm{o1}}\big(e_{\mathrm{s1}}e_{\mathrm{o1}}e_{\mathrm{o2}}  + e_{\mathrm{s2}}e_{\mathrm{o1}}^2 \big) $

\vspace{4mm}
$\mathcal{T}_{\mathrm{s1}}e_{\mathrm{s1}}\big(-e_{\mathrm{s2}}e_{\mathrm{o1}} +e_{\mathrm{s2}}e_{\mathrm{o2}} - e_{\mathrm{o1}}^2 + e_{\mathrm{o1}}e_{\mathrm{o2}}\big) + \mathcal{T}_{\mathrm{o1}}e_{\mathrm{o1}}\big(e_{\mathrm{s2}}e_{\mathrm{o2}}  + e_{\mathrm{s1}}e_{\mathrm{o1}} \big) \neq $

$\mathcal{T}_{\mathrm{s2}}e_{\mathrm{s2}}\big(-e_{\mathrm{s1}}e_{\mathrm{o1}} + e_{\mathrm{s1}}e_{\mathrm{o2}} -e_{\mathrm{o1}}^2 + e_{\mathrm{o1}}e_{\mathrm{o2}}\big) +  \mathcal{T}_{\mathrm{o1}}e_{\mathrm{o1}}\big(e_{\mathrm{s1}}e_{\mathrm{o2}}  + e_{\mathrm{s2}}e_{\mathrm{o1}} \big) $

\vspace{4mm}
$\mathcal{T}_{\mathrm{s1}}e_{\mathrm{s1}}\big(-e_{\mathrm{s2}}e_{\mathrm{o1}} +e_{\mathrm{s2}}e_{\mathrm{o2}} - e_{\mathrm{o1}}^2 + e_{\mathrm{o1}}e_{\mathrm{o2}}\big) + \mathcal{T}_{\mathrm{o1}}e_{\mathrm{o1}}\big(e_{\mathrm{s2}}e_{\mathrm{o2}}  + e_{\mathrm{s1}}e_{\mathrm{o1}}  - e_{\mathrm{s1}}e_{\mathrm{o2}}  - e_{\mathrm{s2}}e_{\mathrm{o1}} \big) \neq$

$\mathcal{T}_{\mathrm{s2}}e_{\mathrm{s2}}\big(-e_{\mathrm{s1}}e_{\mathrm{o1}} + e_{\mathrm{s1}}e_{\mathrm{o2}} -e_{\mathrm{o1}}^2 + e_{\mathrm{o1}}e_{\mathrm{o2}}\big)  $

\vspace{4mm}
$\mathcal{T}_{\mathrm{s1}}e_{\mathrm{s1}}\big(e_{\mathrm{s2}}(e_{\mathrm{o2}} - e_{\mathrm{o1}}) + e_{\mathrm{o1}}(e_{\mathrm{o2}} - e_{\mathrm{o1}})\big) + \mathcal{T}_{\mathrm{o1}}e_{\mathrm{o1}}\big(e_{\mathrm{s2}} - e_{\mathrm{s1}}\big)\big(e_{\mathrm{o2}}  - e_{\mathrm{o1}} \big) \neq$

$\mathcal{T}_{\mathrm{s2}}e_{\mathrm{s2}}\big(e_{\mathrm{s1}}(e_{\mathrm{o2}} - e_{\mathrm{o1}}) + e_{\mathrm{o1}}(e_{\mathrm{o2}} - e_{\mathrm{o1}})\big)  $

\vspace{4mm}
$\mathcal{T}_{\mathrm{s1}}e_{\mathrm{s1}}\big(e_{\mathrm{s2}} + e_{\mathrm{o1}}\big)\big(e_{\mathrm{o2}} - e_{\mathrm{o1}}\big) + \mathcal{T}_{\mathrm{o1}}e_{\mathrm{o1}}\big(e_{\mathrm{s2}} - e_{\mathrm{s1}}\big)\big(e_{\mathrm{o2}}  - e_{\mathrm{o1}} \big) \neq$

$\mathcal{T}_{\mathrm{s2}}e_{\mathrm{s2}}\big(e_{\mathrm{s1}} + e_{\mathrm{o1}}\big)\big(e_{\mathrm{o2}} - e_{\mathrm{o1}}\big)  $

\vspace{4mm}
$\Big(\mathcal{T}_{\mathrm{s1}}e_{\mathrm{s1}}\big(e_{\mathrm{s2}} + e_{\mathrm{o1}}\big) + \mathcal{T}_{\mathrm{o1}}e_{\mathrm{o1}}\big(e_{\mathrm{s2}} - e_{\mathrm{s1}}\big)\Big)\big(e_{\mathrm{o2}}  - e_{\mathrm{o1}} \big) \neq$

$\mathcal{T}_{\mathrm{s2}}e_{\mathrm{s2}}\big(e_{\mathrm{s1}} + e_{\mathrm{o1}}\big)\big(e_{\mathrm{o2}} - e_{\mathrm{o1}}\big)  $}

\vspace{4mm}

\normalsize{Given $e_{\mathrm{o2}} \neq e_{\mathrm{o1}}$, we can drop the $(e_{\mathrm{o2}} - e_{\mathrm{o1}})$ term which results in:}

\vspace{4mm}

\small{$\mathcal{T}_{\mathrm{s1}}e_{\mathrm{s1}}\big(e_{\mathrm{s2}} + e_{\mathrm{o1}}\big) + \mathcal{T}_{\mathrm{o1}}e_{\mathrm{o1}}\big(e_{\mathrm{s2}} - e_{\mathrm{s1}}\big) \neq \mathcal{T}_{\mathrm{s2}}e_{\mathrm{s2}}\big(e_{\mathrm{s1}} + e_{\mathrm{o1}}\big)  $}

\vspace{4mm}




\vspace{4mm}

\normalsize{Rearranging,}

\vspace{4mm}

\small{$ \mathcal{T}_{\mathrm{s1}}e_{\mathrm{s1}}e_{\mathrm{s2}}  + \mathcal{T}_{\mathrm{s1}}e_{\mathrm{s1}}e_{\mathrm{o1}} -\mathcal{T}_{\mathrm{s2}}e_{\mathrm{s1}}e_{\mathrm{s2}} - \mathcal{T}_{\mathrm{s2}}e_{\mathrm{s2}}e_{\mathrm{o1}} - \mathcal{T}_{\mathrm{o1}}e_{\mathrm{s1}}e_{\mathrm{o1}} + \mathcal{T}_{\mathrm{o1}}e_{\mathrm{s2}}e_{\mathrm{o1}} \neq 0$

\vspace{4mm}

$e_{\mathrm{s1}}e_{\mathrm{s2}}\big( \mathcal{T}_{\mathrm{s1}} -  \mathcal{T}_{\mathrm{s2}}\big)  + e_{\mathrm{s1}}e_{\mathrm{o1}}\big(\mathcal{T}_{\mathrm{s1}} - \mathcal{T}_{\mathrm{o1}} \big)   + e_{\mathrm{s2}}e_{\mathrm{o1}}\big( \mathcal{T}_{\mathrm{o1}}  - \mathcal{T}_{\mathrm{s2}}\big) \neq 0$}

\vspace{4mm}

\normalsize{We note that material effusivities are greater than $0$, and sensor temperature is controllable. Thus, we can always choose sensors such that for $e_{\mathrm{s1}} > e_{\mathrm{s2}}$, $\mathcal{T}_{\mathrm{s1}} > \mathcal{T}_{\mathrm{s2}}$. Specifically in our experiments, as one sensor is active, we can assume it will always have a higher temperature than the passive sensor, thus $\mathcal{T}_{\mathrm{s1}} > \mathcal{T}_{\mathrm{s2}}$. Thus, }

\vspace{4mm}

\small{$e_{\mathrm{s1}}e_{\mathrm{s2}}\big( \mathcal{T}_{\mathrm{s1}} -  \mathcal{T}_{\mathrm{s2}}\big)  + e_{\mathrm{o1}}\Big( e_{\mathrm{s1}}\big(\mathcal{T}_{\mathrm{s1}} - \mathcal{T}_{\mathrm{o1}} \big)   + e_{\mathrm{s2}}\big( \mathcal{T}_{\mathrm{o1}}  - \mathcal{T}_{\mathrm{s2}}\big)\Big) > 0$}

\vspace{4mm}

\normalsize{because $ \mathcal{T}_{\mathrm{s1}} -  \mathcal{T}_{\mathrm{s2}} > 0$ and $ e_{\mathrm{s1}}\big(\mathcal{T}_{\mathrm{s1}} - \mathcal{T}_{\mathrm{o1}} \big)   > \big| -e_{\mathrm{s2}}\big( \mathcal{T}_{\mathrm{s2}} - \mathcal{T}_{\mathrm{o1}}\big)\big|$.}

\vspace{4mm}

We also note a special condition that occurs when the passive and active sensor have the same effusivity, i.e. $e_{\mathrm{s1}} = e_{\mathrm{s2}} = e_{\mathrm{s}}$. Starting from Eq.~4, with two sensors with initial temperatures $\mathcal{T}_{\mathrm{s1}}$ and $\mathcal{T}_{\mathrm{s2}}$ and effusivities $e_{\mathrm{s}}$,

\vspace{4mm}

\setlength{\parindent}{0cm}
\large{${-\mathcal{T}_{\mathrm{s1}}e_{\mathrm{s}}e_{\mathrm{o1}} + \mathcal{T}_{\mathrm{s1}}e_{\mathrm{s}}e_{\mathrm{o2}} + \mathcal{T}_{\mathrm{o1}}e_{\mathrm{s}}e_{\mathrm{o1}} + \mathcal{T}_{\mathrm{o1}}e_{\mathrm{o1}}e_{\mathrm{o2}} \over e_{\mathrm{o2}}(e_{\mathrm{s}} + e_{\mathrm{o1}})} \neq {-\mathcal{T}_{\mathrm{s2}}e_{\mathrm{s}}e_{\mathrm{o1}} + \mathcal{T}_{\mathrm{s2}}e_{\mathrm{s}}e_{\mathrm{o2}} + \mathcal{T}_{\mathrm{o1}}e_{\mathrm{s}}e_{\mathrm{o1}} + \mathcal{T}_{\mathrm{o1}}e_{\mathrm{o1}}e_{\mathrm{o2}} \over e_{\mathrm{o2}}(e_{\mathrm{s}} + e_{\mathrm{o1}})} $}

\vspace{4mm}

\small{$-\mathcal{T}_{\mathrm{s1}}e_{\mathrm{s}}e_{\mathrm{o1}} + \mathcal{T}_{\mathrm{s1}}e_{\mathrm{s}}e_{\mathrm{o2}} + \mathcal{T}_{\mathrm{o1}}e_{\mathrm{s}}e_{\mathrm{o1}} + \mathcal{T}_{\mathrm{o1}}e_{\mathrm{o1}}e_{\mathrm{o2}} \neq -\mathcal{T}_{\mathrm{s2}}e_{\mathrm{s}}e_{\mathrm{o1}} + \mathcal{T}_{\mathrm{s2}}e_{\mathrm{s}}e_{\mathrm{o2}} + \mathcal{T}_{\mathrm{o1}}e_{\mathrm{s}}e_{\mathrm{o1}} + \mathcal{T}_{\mathrm{o1}}e_{\mathrm{o1}}e_{\mathrm{o2}}$}

\vspace{4mm}

\small{$-\mathcal{T}_{\mathrm{s1}}e_{\mathrm{s}}e_{\mathrm{o1}} + \mathcal{T}_{\mathrm{s1}}e_{\mathrm{s}}e_{\mathrm{o2}} \neq -\mathcal{T}_{\mathrm{s2}}e_{\mathrm{s}}e_{\mathrm{o1}} + \mathcal{T}_{\mathrm{s2}}e_{\mathrm{s}}e_{\mathrm{o2}} $}

\vspace{4mm}

\small{$-\mathcal{T}_{\mathrm{s1}}e_{\mathrm{o1}} + \mathcal{T}_{\mathrm{s1}}e_{\mathrm{o2}} \neq -\mathcal{T}_{\mathrm{s2}}e_{\mathrm{o1}} + \mathcal{T}_{\mathrm{s2}}e_{\mathrm{o2}} $}

\vspace{4mm}

\normalsize{$e_{\mathrm{s}}$ is strictly greater than $0$, so,}

\vspace{4mm}

\small{$-\mathcal{T}_{\mathrm{s1}}e_{\mathrm{o1}} + \mathcal{T}_{\mathrm{s1}}e_{\mathrm{o2}} \neq -\mathcal{T}_{\mathrm{s2}}e_{\mathrm{o1}} + \mathcal{T}_{\mathrm{s2}}e_{\mathrm{o2}} $}

\vspace{4mm}

\small{$\mathcal{T}_{\mathrm{s1}}\big(e_{\mathrm{o2}} - e_{\mathrm{o1}}\big) \neq \mathcal{T}_{\mathrm{s2}}\big(e_{\mathrm{o2}} -e_{\mathrm{o1}}\big) $}

\vspace{4mm}

\normalsize{which holds given $e_{\mathrm{o2}} \neq e_{\mathrm{o1}}$}.

\begin{figure}
\begin{center}
\includegraphics[width=8.7cm]{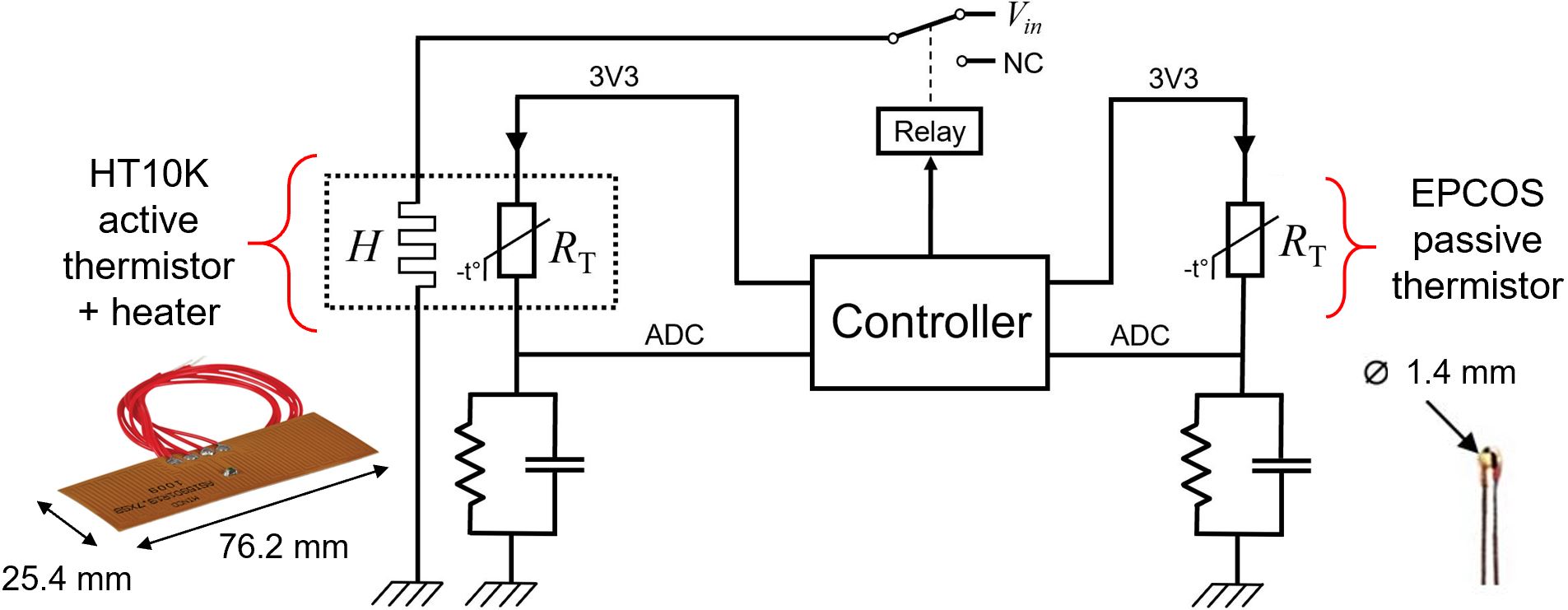} 
\vspace{-2mm}
\caption{\label{fig:8}\textbf{Thermal Sensing Equipment and Circuitry.} We used a Teensy 3.2 on the robot, coupled to a heated thermistor for active thermal sensing and an non-heated thermistor for passive sensing. The HT10K used for active thermal sensing is folded twice to form a square with 3 layers and mounted on the end of the robot. The robot is controlled to make contact with the material when the heater ($\mathcal{T}_{\mathrm{r}}$) reaches the mean human finger temperature measured in the first session ($\mathcal{T}_{\mathrm{r}} = \overline{\mathcal{T}}_{\mathrm{f}} = 29.5^\circ\mathrm{C}$). The system uses a voltage divider circuit and the internal analog-to-digital converter in the microcontroller to measure the resistance of the thermistors. }
\end{center}
\end{figure}

\section{Robot Circuitry}

We used a Teensy 3.2 microcontroller to control the Firgelli L12 linear actuator and read data from the sensor module on the robot's end-effector, as well as the separate passive thermistor mounted to the cold wood material sample during the experiment. Fig. \ref{fig:8} shows the circuit used for combined active and passive thermal sensing. The left side shows a circuit diagram of the active sensing module that uses a Thorlabs HT10K combined thermistor and heat source. The passive circuit on the right portion contains an EPCOS B57541G1103F thermistor. For temperature, we measured the thermistor resistance using a voltage divider circuit and the microcontroller's internal 12-bit analog-to-digital converter. This resistance value is then converted to temperature in Celsius using calibration constants provided by the thermistor manufacturer. 
Both thermistors in the circuit have the same nominal resistance ($10\,\mathrm{k\Omega}$), a reference resistor of ($10\,\mathrm{k\Omega}$) to limit heat generation, and a constant $3.3\,\mathrm{V}$ input from the microcontroller. For both the active and passive thermistors we included a physical $.022\,\mathrm{\mu F}$ capacitor in parallel with the reference resistor and 2nd order software Butterworth filter with a cutoff frequency of $10\,Hz$ to filter noise. We control the robot to touch a material when the active sensor reaches a some specified temperature.

\ifCLASSOPTIONcaptionsoff
  \newpage
\fi



\bibliographystyle{IEEEtran}
%
